\documentclass[journal]{IEEEtran}
\ifCLASSINFOpdf
\else
\fi

\usepackage{times}
\usepackage{epsfig}
\usepackage{graphicx}
\usepackage{amsmath}
\usepackage{amssymb}
\usepackage{subfig}
\usepackage{bm}
\usepackage{multirow}  
\usepackage{booktabs}
\usepackage{comment}

\usepackage[table]{xcolor}
\usepackage{arydshln}
\definecolor{mygray}{gray}{.92}

\makeatletter
\newcommand{\thickhline}{%
    \noalign {\ifnum 0=`}\fi \hrule height 1pt
    \futurelet \reserved@a \@xhline
}

\usepackage[pagebackref=true,breaklinks=true,colorlinks,bookmarks=false,citecolor=green]{hyperref}

\begin{document}
%
\title{Efficient Person Search: An Anchor-Free Approach}
%
%
%

\author{Yichao Yan,
        Jinpeng Li,
        Jie Qin,~\IEEEmembership{Member,~IEEE},
        Shengcai Liao,~\IEEEmembership{Senior Member,~IEEE},
        and {Xiaokang Yang},~\IEEEmembership{Fellow,~IEEE}
\thanks{Y. Yan and X. Yang are with MoE Key Lab of Artificial Intelligence, AI Institute, Shanghai Jiao Tong University, Shanghai, China. E-mail: \{yanyichao, xkyang\}@sjtu.edu.cn.} 
\thanks{J. Li, J. Qin and S. Liao are with the Inception Institute of Artificial Intelligence, Abu Dhabi, United Arab Emirates. E-mail: ljpadam@gmail.com, qinjiebuaa@gmail.com, scliao@ieee.org.}

}

\maketitle

\begin{abstract}
Person search aims to simultaneously localize and identify a query person from realistic, uncropped images. To achieve this goal, state-of-the-art models typically add a re-id branch upon two-stage detectors like Faster R-CNN. Owing to the ROI-Align operation, this pipeline yields promising accuracy as re-id features are explicitly aligned with the corresponding object regions, but in the meantime, it introduces high computational overhead due to dense object anchors. In this work, we present an anchor-free approach to efficiently tackling this challenging task, by introducing the following dedicated designs. \textit{First}, we select an anchor-free detector (i.e., FCOS) as the prototype of our framework. Due to the lack of dense object anchors, it exhibits significantly higher efficiency compared with existing person search models.
\textit{Second}, when directly accommodating this anchor-free detector for person search, there exist several major challenges in learning robust re-id features, which we summarize as the misalignment issues in different levels (i.e., scale, region, and task). To address these issues, we propose an aligned feature aggregation module to generate more discriminative and robust feature embeddings. Accordingly, we name our model as Feature-Aligned Person Search Network (AlignPS).
\textit{Third}, by investigating the advantages of both anchor-based and anchor-free models, we further augment AlignPS with an ROI-Align head, which significantly improves the robustness of re-id features while still keeping our model highly efficient. 
Extensive experiments conducted on two challenging benchmarks (i.e., CUHK-SYSU and PRW) demonstrate that our framework achieves state-of-the-art or competitive performance, while displaying higher efficiency. All the source codes, data, and trained models are available at: \url{ https://github.com/daodaofr/alignps}.

\end{abstract}


%
\IEEEpeerreviewmaketitle

\section{Introduction}


\IEEEPARstart{P}{erson} search~\cite{DBLP:conf/cvpr/ZhengZSCYT17,DBLP:conf/cvpr/XiaoLWLW17} aims to localize and identify a target person from a gallery of realistic, uncropped scene images, and it has recently emerged as a practical task with real-world applications, e.g., video surveillance. Two fundamental computer vision tasks, i.e., pedestrian detection~\cite{DBLP:conf/iccv/OuyangW13,DBLP:conf/cvpr/ZhangBS17} and person re-identification (re-id)~\cite{DBLP:conf/cvpr/FarenzenaBPMC10,DBLP:conf/cvpr/AhmedJM15}, need to be addressed to tackle this task. Both detection and re-id are very challenging tasks and have received tremendous attention in the past decade. In person search, we need to not only address the challenges (e.g., occlusions, pose/viewpoint variations, and background clutter) of the two individual tasks, but also pursue a unified and optimized framework to simultaneously perform detection and re-id.

\begin{figure*}[t]
\centering
\subfloat[Two-step person search framework\label{subfig-1-1}]{%
   \includegraphics[width=0.45\linewidth]{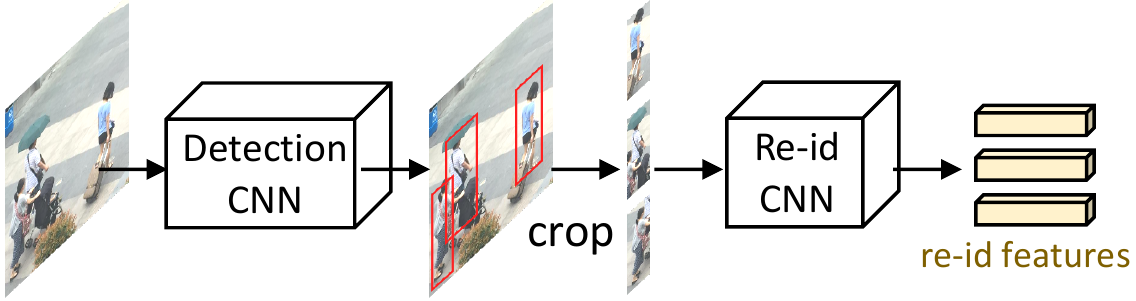}
}
\hspace{5mm}
\subfloat[One-step two-stage person search framework\label{subfig-1-2}]{%
   \includegraphics[width=0.45\linewidth]{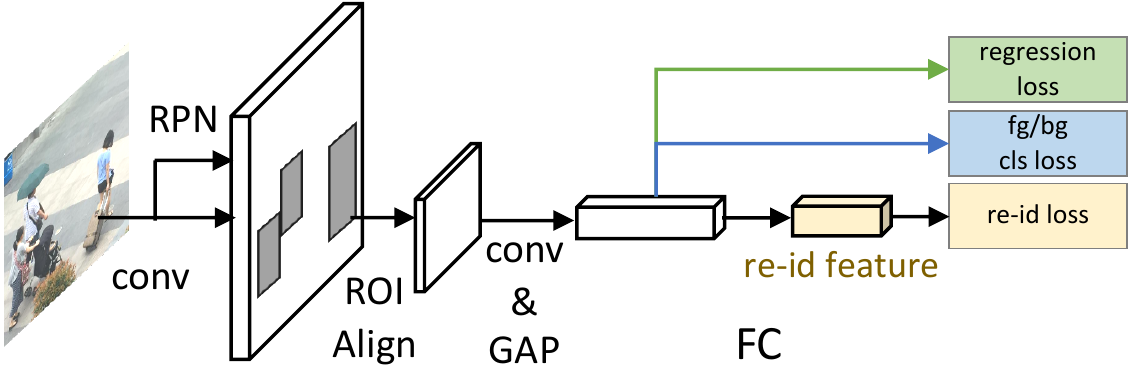}
}
\hspace{1mm}
\subfloat[AlignPS: the proposed one-step one-stage anchor-free framework\label{subfig-1-3}]{%
   \includegraphics[width=0.45\linewidth]{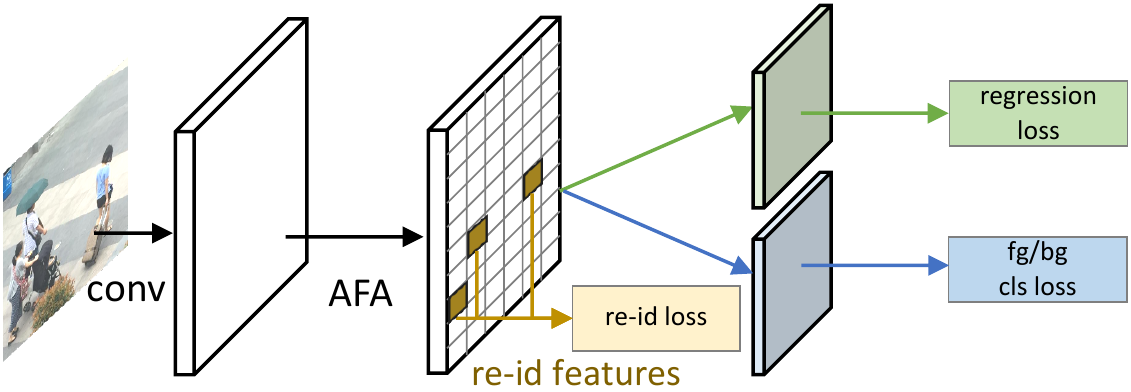}
}
\hspace{5mm}
\subfloat[ROI-AlignPS: AlignPS augmented with an ROI-Align head\label{subfig-1-4}]{%
   \includegraphics[width=0.45\linewidth]{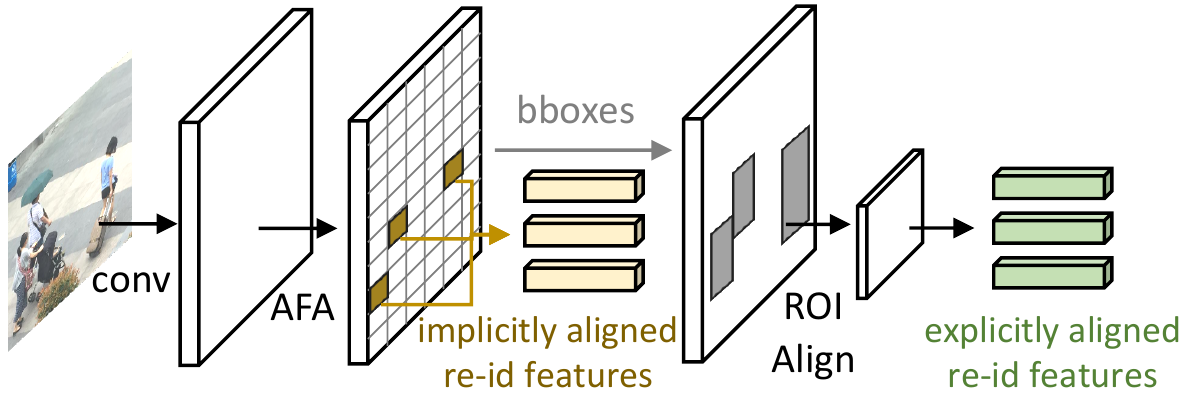}
}
 \caption{Comparison of four person search frameworks. (a) The two-step framework addresses detection and re-id as two separate tasks. (b) The one-step model enables end-to-end training of detection and re-id with an ROI-Align operation based on a two-stage detector; however, re-id is considered as a secondary task after detection. (c) The proposed framework enables single-stage inference for both detection and re-id, while making re-id the primary task. (d) By appending an ROI-Align head to the bounding box output of AlignPS, the re-id features are further improved with explicit region alignment. }
 \label{fig:intro}
\end{figure*}

Previous person search frameworks can be generally divided into two categories. The first line of works~\cite{DBLP:conf/cvpr/ZhengZSCYT17,DBLP:conf/eccv/ChenZOYT18,DBLP:conf/eccv/LanZG18} can be summarized as \emph{two-step} approaches, which attempt to deal with detection and re-id separately. As shown in Fig.~\ref{subfig-1-1}, multiple persons are first localized with off-the-shelf detection models, and then cropped out and fed to re-id networks to extract feature representations. Although two-step models can obtain satisfactory results, the disentangled treatment of the two tasks is time- and resource-consuming.
In contrast, the second category~\cite{DBLP:conf/cvpr/XiaoLWLW17,DBLP:conf/iccv/LiuFJKZQJY17,DBLP:conf/eccv/ChangHSLYH18,DBLP:conf/cvpr/MunjalATG19,DBLP:conf/cvpr/ChenZYS20} provides a \emph{one-step} solution that unifies detection and re-id in an end-to-end manner.
As shown in Fig.~\ref{subfig-1-2}, one-step models first apply an ROI-Align layer to aggregate features in the detected bounding boxes. The features are then shared by detection and re-id; with an additional re-id loss, the simultaneous optimization of the two tasks becomes feasible. Since these models adopt two-stage detectors like Faster R-CNN~\cite{DBLP:journals/pami/RenHG017}, we refer to them as \emph{one-step two-stage} models. However, these methods inevitably inherit the limitations of two-stage detectors, e.g., high computational complexity caused by dense anchors, and high sensitivity to the hyperparameters including the size, aspect ratio and number of anchor boxes, etc.

Compared with two-stage detectors, anchor-free models exhibit unique advantages (e.g., simpler structure and higher speed), and have been actively studied in recent years \cite{DBLP:conf/cvpr/RedmonDGF16,DBLP:conf/eccv/LawD18,DBLP:conf/cvpr/LiuLRHY19,DBLP:conf/iccv/DuanBXQH019}. Inspired by this, an open question is naturally thrown at us - \emph{Is it possible to develop an anchor-free framework for person search?} Our answer is yes. However, this is a non-trivial task due to the following three misalignment issues. \textbf{1)} Many anchor-free models learn multi-scale features using feature pyramid networks (FPNs) \cite{DBLP:conf/cvpr/LinDGHHB17} to achieve scale invariance for object detection. However, this introduces the misalignment issue for re-id (i.e., scale misalignment), as a query person needs to be compared with all the people of various scales in the gallery set, while the re-id features would be inconsistently taken from different FPN levels. \textbf{2)} In the absence of operations like ROI-Align, anchor-free models cannot align the features for re-id and detection according to a specific region. Therefore, re-id embeddings must be directly learned from feature maps without explicit region alignment, which brings additional challenges as re-id features are sensitive to the foreground regions~\cite{DBLP:conf/cvpr/DongZST20a,DBLP:conf/iccv/HanYZTZGS19}. \textbf{3)} Person search can be intuitively formulated as a multi-task learning framework with detection and re-id as its sub-tasks. However, there exist conflicts between the objectives of these two tasks, i.e., pedestrian detection tries to learn features that are commonly shared by all the people, while re-id tries to extract unique features for individual persons. Hence, we need to find a better tradeoff/alignment between the two tasks.

In this work, we present an anchor-free framework for efficient person search, which we name the Feature-Aligned Person Search Network (\textbf{AlignPS}). Our model employs the typical architecture of an anchor-free detection model (i.e., FCOS~\cite{DBLP:conf/iccv/TianSCH19}), which allows our framework to be more efficient than prior person search models. 
To address the above-mentioned challenges, we design an aligned feature aggregation (AFA) module to make our model focus more on the re-id subtask. Specifically, AFA reshapes some building blocks of FPN to overcome the issues of region and scale misalignment in re-id feature learning. For example, we exploit deformable convolution to make the re-id embeddings adaptively aligned with the foreground regions. In the meantime, we design a feature fusion scheme to better aggregate features from different FPN levels, which makes the re-id features more robust to scale variations. We also optimize the training procedures of re-id and detection to place more emphasis on generating robust re-id embeddings (as shown in Fig.~\ref{subfig-1-3}). These simple yet effective designs successfully transform a classic anchor-free detector into a powerful and efficient person search framework, and allow the proposed model to outperform its anchor-based competitors.
Moreover, we find that although the proposed AlignPS framework implicitly aligns re-id features with the corresponding regions, there inevitably exists a gap between the adapted regions and the foreground bounding boxes. As observed in previous works~\cite{DBLP:conf/cvpr/DongZST20a}, re-id features are sensitive to the context outside the foreground regions. Inspired by the learning scheme in two-stage models (as shown in Fig.~\ref{subfig-1-2}), we further augment AlignPS with an ROI-Align head, to explicitly extract more robust re-id features corresponding to the foreground regions. We name this variant as \textbf{ROI-AlignPS}, as shown in Fig.~\ref{subfig-1-4}. These explicitly aligned re-id features can be viewed as complements to the implicitly aligned features from AlignPS, and these two kinds of features are fused to yield even better re-id representations. More importantly, the ROI-Align head only receives the output bounding boxes from AlignPS during inference, avoiding computing dense region proposals. Therefore, ROI-AlignPS still inherits the high efficiency from AlignPS.


In summary, our main contributions include:
\begin{itemize}
\setlength{\itemsep}{0pt}
	\setlength{\parsep}{-3pt}
	\setlength{\parskip}{-0pt}
	\setlength{\leftmargin}{-15pt}
    \item We propose the first \emph{anchor-free} framework (AlignPS) for efficient person search, which will significantly foster future research in this direction.
    \item We design an AFA module that simultaneously addresses the issues of scale, region, and task misalignment to successfully accommodate an anchor-free detector for the task of person search.
    \item We further propose a novel model variant (ROI-AlignPS) by augmenting AlignPS with an ROI-Align head, which complements the implicitly aligned re-id features. With mutual learning strategy, this variant further improves the performance while remaining highly efficient.
    \item As an anchor-free framework, our model surprisingly achieves state-of-the-art or competitive performance on two challenging person search benchmarks, while running at a higher speed.
\end{itemize}

Part of this work has been published in \cite{Yan_2021_CVPR}. In this paper, we further make the following extensions: \textbf{1)} We augment AlignPS with an ROI-Align head, which explicitly aligns the foreground regions with their re-id features. This architecture variant notably improves the performance, while keeping the framework efficient. \textbf{2)} We investigate several mutual feature learning strategies, which allow the explicitly and implicitly aligned re-id features to promote each other during training. We find these strategies yield more robust re-id representations. \textbf{3)} We provide more thorough ablation studies and component analysis on both CUHK-SYSU and PRW, and present more qualitative results. These analyses further illustrate the effectiveness of the proposed framework.


\section{Related Work}
\textbf{Pedestrian Detection}. Pedestrian or object detection can be considered as a preliminary task of person search. Current deep learning-based detectors are generally categorized into one-stage and two-stage models, according to whether they employ a region proposal network (RPN) to generate object proposals. Alternatively, object detectors can also be categorized into anchor-based and anchor-free detectors, depending on whether they utilize anchor boxes to associate objects. One of the most representative two-stage anchor-based detectors is Faster R-CNN~\cite{DBLP:journals/pami/RenHG017}, which has been extended into numerous variants~\cite{DBLP:conf/iccv/DaiQXLZHW17,DBLP:conf/cvpr/CaiV18,DBLP:conf/cvpr/PangCSFOL19,DBLP:conf/cvpr/SongLW20}. Notably, some one-stage detectors~\cite{DBLP:conf/eccv/LiuAESRFB16,DBLP:conf/iccv/LinGGHD17,DBLP:conf/cvpr/RedmonF17,DBLP:conf/cvpr/ZhangWBLL18} also work with anchor boxes. Compared with the above models, one-stage anchor-free detectors~\cite{DBLP:conf/cvpr/RedmonDGF16,DBLP:conf/eccv/LawD18,DBLP:conf/cvpr/LiuLRHY19,DBLP:journals/corr/abs-1904-07850,DBLP:conf/iccv/YangLHWL19,DBLP:conf/iccv/TianSCH19,DBLP:conf/mm/LiLJ020} have been attracting more and more attention recently due to their simple structures and 
efficient implementations. In this work, we develop our person search framework based on a classic one-stage anchor-free detector, thus making the whole framework simpler and faster.    

\textbf{Person Re-identification}. Person re-id is also closely related to person search, aiming to learn identity embeddings from cropped person images. Traditional methods employed various handcrafted features~\cite{DBLP:journals/ijcv/Lowe04,DBLP:conf/cvpr/FarenzenaBPMC10,DBLP:conf/eccv/GrayT08} before the renaissance of deep learning. However, to pursue better performance, current re-id models are mostly based on deep learning. Some models employ structure/part information in the human body to learn more robust representations~\cite{DBLP:conf/iccv/SuLZX0T17,DBLP:conf/eccv/SunZYTW18,DBLP:conf/iccv/MiaoWLD019,9233968}, while others focus on learning better distance metrics~\cite{DBLP:conf/cvpr/AhmedJM15,DBLP:journals/corr/HermansBL17,DBLP:conf/cvpr/ChenCZH17,DBLP:journals/pami/ChenZZL18,DBLP:journals/pami/WangGZW16,DBLP:journals/tip/ZhuWHZ18}. As person re-id usually lacks large-scale training data, data augmentation~\cite{DBLP:conf/nips/GeLZYYWL18,DBLP:conf/cvpr/LiuNYZCH18,DBLP:conf/cvpr/WeiZ0018,DBLP:journals/tip/ZhongZZLY19} also becomes popular for tackling this task. Compared with detection which aims to learn common features of pedestrians, re-id needs to focus more on fine-grained details and unique features of each identity. Therefore, we propose to follow the ``re-id first" principle to raise the priority of the re-id task, resulting in more discriminative identity embeddings for more accurate person search.

\begin{figure*}[t]
\setlength{\abovecaptionskip}{1mm}
\centering
\includegraphics[width=\linewidth]{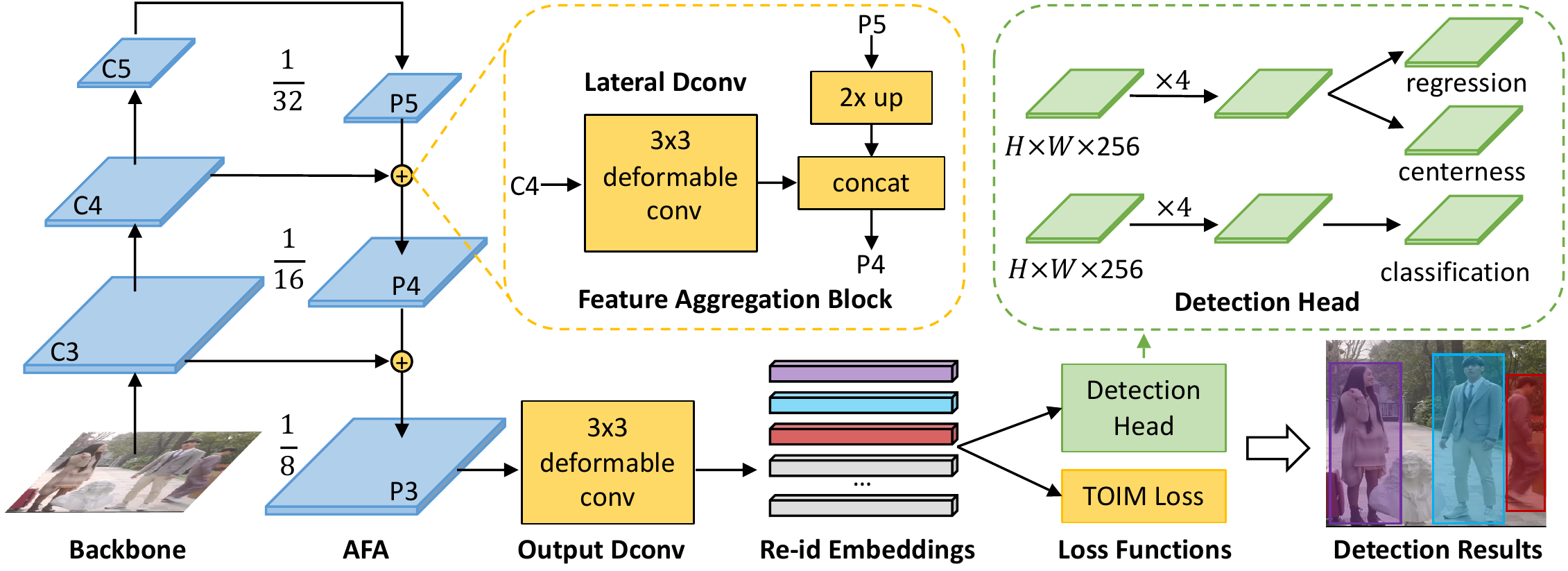}
\caption{Architecture of the proposed AlignPS framework, which shares the basic structure of FCOS~\cite{DBLP:conf/iccv/TianSCH19}. The components in yellow are newly designed to accommodate FCOS for the task of person search. ``Dconv'' means deformable convolution.}
\label{fig:arch}
\vspace{-2mm}
\end{figure*}

\textbf{Person Search}. Existing person search frameworks can be divided into two-step and one-step models. Two-step models first perform pedestrian detection and subsequently crop the detected people for re-id. Zheng et al~\cite{DBLP:conf/cvpr/ZhengZSCYT17} introduced the first two-step framework for person search and evaluated the combinations of different detectors and re-id models. Since then, several models~\cite{DBLP:conf/eccv/ChenZOYT18,DBLP:conf/eccv/LanZG18,DBLP:conf/iccv/HanYZTZGS19,DBLP:conf/cvpr/WangMCSC20} have followed this pipeline. In~\cite{DBLP:conf/cvpr/XiaoLWLW17}, Xiao et al. proposed the first one-step person search framework based on Faster R-CNN. Specifically, a joint framework enabling end-to-end training of detection and re-id was proposed by stacking a re-id embedding layer after the detection features and proposing the Online Instance Matching (OIM) loss. So far, a number of improvements~\cite{DBLP:conf/iccv/LiuFJKZQJY17,DBLP:journals/pr/XiaoXTHWF19,DBLP:conf/eccv/ChangHSLYH18,DBLP:conf/cvpr/YanZNZXY19,DBLP:conf/cvpr/MunjalATG19,DBLP:conf/cvpr/DongZST20a,DBLP:conf/cvpr/ChenZYS20} have been made based on this framework. In general, two-step models may achieve better performance, while one-step models have the advantages of simplicity and efficiency. However, there is still room for improving one-step methods due to the aforementioned shortcomings of the two-stage anchor-based detectors they usually adopt. In this work, we introduce the first anchor-free model to further improve the simplicity and efficiency of one-step models, without any sacrifice in accuracy. 

\textbf{Mutual Learning}. Hinton et al~\cite{DBLP:journals/corr/HintonVD15} firstly introduced knowledge distillation in neural networks in 2015. Since then, it has been widely employed in various computer vision tasks to improve the capability of neural networks, e.g., image recognition~\cite{DBLP:journals/pami/LiH18a,DBLP:conf/eccv/WangFLWLM20,DBLP:conf/iccv/PengLZLQT19}, object detection~\cite{DBLP:conf/cvpr/LiJY17,DBLP:conf/nips/ChenCYHC17,DBLP:conf/cvpr/ZhuHLD19}, image segmentation~\cite{DBLP:conf/iccv/MullapudiCZRF19,DBLP:conf/cvpr/LiuCLQLW19,DBLP:conf/cvpr/Hou0LHL20}, etc. In person search, several recent work have also adopted this approach. For example, QEEPS~\cite{DBLP:conf/cvpr/MunjalATG19} and IGPN~\cite{DBLP:conf/cvpr/DongZST20} employ the query person to find the most similar proposals in the RPN, which reduces the number of proposals and enhances the efficiency. In BiNet~\cite{DBLP:conf/cvpr/DongZST20a} and DKD~\cite{DBLP:conf/aaai/ZhangWBSY21}, the features from cropped pedestrians are utilized to distill the re-id features in person search network, such that the robustness of the re-id features are improved. In this work, we design two feature learning branches in ROI-AlignPS. Instead of employing the query person or the cropped image, we investigate several mutual feature promotion strategies to improve the discriminative capability of the aggregated features.

\section{Feature-Aligned Person Search Networks}
In this section, we introduce the proposed anchor-free frameworks (i.e., AlignPS and ROI-AlignPS) for person search. First, we give an overview of the network architecture. Second, the proposed AFA module is elaborated with the aim of mitigating different levels of misalignment issues when transforming an anchor-free detector into a superior person search framework. Then, we present the designed loss function to obtain more discriminative features for person search. Finally, we present ROI-AlignPS, a model variant which further improves the performance of AlignPS.

\subsection{Framework Overview}
The basic framework of the proposed AlignPS is based on FCOS~\cite{DBLP:conf/iccv/TianSCH19}, one of the most popular one-stage anchor-free object detectors. Differently, we adhere to the ``re-id first" principle to put emphasis on learning robust feature embeddings for the re-id subtask, which is crucial for enhancing the overall performance of person search.

As illustrated in Fig.~\ref{fig:arch}, our model simultaneously localizes multiple people in the image and learns re-id embeddings for them. Specifically, an AFA module is developed to aggregate features from multi-level feature maps in the backbone network. To learn re-id embeddings, which is the key of our method, we directly take the flattened features from the output feature maps of AFA as the final embeddings, without any extra embedding layers. For detection, we employ the detection head from FCOS which is good enough for the detection subtask. The detection head consists of two branches, both of which contain four 3$\times$3 \emph{conv} layers. In the meantime, the first branch predicts regression offsets and centerness scores, while the second makes foreground/background classification. Finally, each location on the output feature map of AFA will be associated with a bounding box with classification and centerness scores, as well as a re-id feature embedding.

\subsection{Aligned Feature Aggregation}
Following FPN~\cite{DBLP:conf/cvpr/LinDGHHB17}, we make use of different levels of feature maps to learn detection and re-id features. As the key of our framework, the proposed AFA performs three levels of alignment, beyond the original FPN, to make the output re-id features more discriminative.

\begin{figure}[t]
\centering
\includegraphics[width=\linewidth]{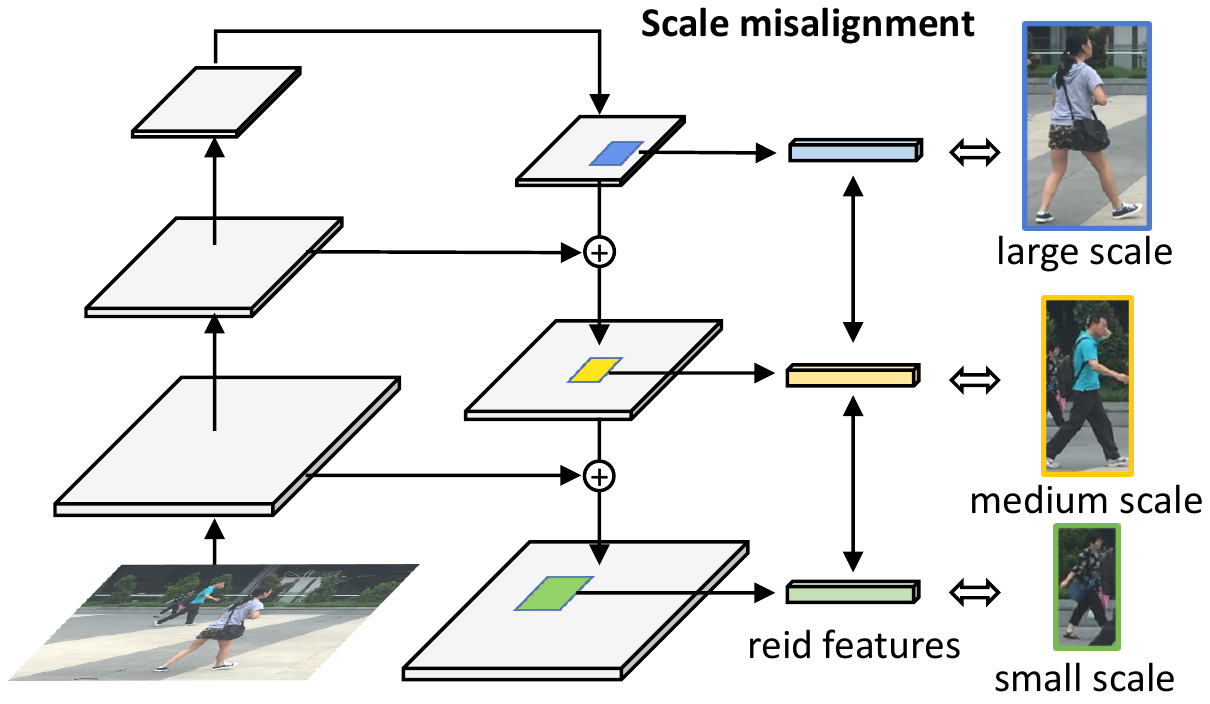}
\caption{Illustration of scale misalignment.}
\label{fig:sm}
\end{figure}

\textbf{Scale Alignment}. As shown in Fig.~\ref{fig:sm}, the original FCOS model employs different levels of features to detect objects of different sizes. This significantly improves the detection performance since the overlapped ambiguous samples will be assigned to different layers. For the re-id task, however, the multi-level prediction could cause feature misalignment between different scales.
In other words, when matching a person of different scales, re-id features are inconsistently taken from different levels of FPN, thus preventing the re-id features from being robust to scale variations.
Furthermore, the people in the gallery set are of various scales, which could eventually make the multi-level model fail to find correct matches for the query person. Therefore, in our framework, we only make predictions based on a single layer of AFA, which explicitly addresses the feature misalignment caused by scale variations. Specifically, we employ the $\{C_3, C_4, C_5\}$ feature maps from the ResNet-50 backbone, and AFA sequentially outputs $\{P_5, P_4, P_3\}$, with strides of 32, 16, and 8, respectively. We only generate features from $\{P_3\}$, which is the largest output feature map, for both the detection and re-id subtasks, and $\{P_6, P_7\}$ are no longer generated as in the original FPN. Although this design may slightly influence the detection performance, we will show in Sec.~\ref{sec:analytical} that it achieves a good trade-off between the detection and re-id subtasks, which adapts well to the person search task.

\textbf{Region Alignment}. On the output feature map of AFA, each location perceives the information from the whole input image based on a large receptive field. Due to the lack of the ROI-Align operation as in Faster R-CNN, it is difficult for our anchor-free framework to learn more accurate features within the pedestrian bounding boxes, and thus leading to the issue of region misalignment. The re-id subtask is even more sensitive to this issue as background features could greatly impact the discriminative capability of the learned features. In AlignPS, we address this issue from three perspectives. \emph{First}, we replace the 1$\times$1 \emph{conv} layers in the lateral connections with 3$\times$3 \emph{deformable conv} layers. As the original lateral connections are designed to reduce the channels of feature maps, a 1$\times$1 \emph{conv} is enough. In our design, moreover, the 3$\times$3 \emph{deformable conv} enables the network to adaptively adjust the receptive field on the input feature maps, thus implicitly fulfilling region alignment. \emph{Second}, we replace the ``sum" operation in the top-down pathway with a ``concatenation" operation, which can better aggregate multi-level features. \emph{Third}, we again replace the 3$\times$3 \emph{conv} with a 3$\times$3 \emph{deformable conv} for the output layer of FPN, which further aligns the multi-level features to finally generate a more accurate feature map. The above three designs work seamlessly to address the region misalignment issue, and we notice that these simple designs are extremely effective when accommodating the basic anchor-free model for our person search task.

\textbf{Task Alignment}. Existing person search frameworks typically treat pedestrian detection as the primary task, i.e., re-id embeddings are just generated by stacking an additional layer after the detection features. A recent work~\cite{DBLP:journals/corr/abs-2004-01888} investigated a parallel structure by employing independent heads for the two tasks to achieve robust multiple object tracking results. In our task of person search, we find the inferior re-id features largely hinder the overall performance. Therefore, we opt for a different principle to align these two tasks by treating re-id as our primary task. Specifically, the output features of AFA are directly supervised with a re-id loss (which will be introduced in the following subsection), and then fed to the detection head. This ``re-id first" design is based on two considerations. \emph{First}, the detection subtask has been relatively well addressed by existing person search frameworks, which directly inherit the advantages from existing powerful detection frameworks. Therefore, learning discriminative re-id embeddings is our primary concern. As we discussed, re-id performance is more sensitive to region misalignment in an anchor-free framework. Therefore, 
it is desirable for the person search framework to be inclined towards the re-id subtask. We also show in our experiments that this design significantly improves the discriminative capability of the re-id embeddings, while having negligible impact on detection. \emph{Second}, compared with ``detection first" and parallel structures, the proposed ``re-id first" structure does not require an extra layer to generate re-id embeddings, and is thus more efficient.

\subsection{Triplet-Aided Online Instance Matching Loss}
Existing works typically employ the OIM loss to supervise the training of the re-id subtask. Specifically, OIM stores the feature centers of all labeled identities in a lookup table (LUT), $V \in \mathbb{R}^{D \times L} = \{v_1,...,v_L\}$, which contains $L$ feature vectors with $D$ dimensions. Meanwhile, a circular queue $U \in \mathbb{R}^{D \times Q}= \{u_1,...,u_Q\}$ containing the features of $Q$ unlabeled identities is maintained. At each iteration, given an input feature $x$ with label $i$, OIM computes the similarity between $x$ and all the features in the LUT and circular queue by $V^T x$ and $Q^T x$, respectively. The probability of $x$ belonging to the identity $i$ is calculated as:
\begin{equation}
    p_i = \frac{{\rm exp}(v_i^T x) / \tau}{\sum_{j=1}^{L}{\rm exp}(v_j^T x) / \tau + \sum_{k=1}^{Q}{\rm exp}(u_k^T x) / \tau}, \label{eq:one}
\end{equation}
where $\tau=0.1$ is a hyperparameter that controls the softness of the probability distribution. The objective of OIM is to minimize the expected negative log-likelihood:
\begin{equation}
    \mathcal{L}_{\text{OIM}} = -{\rm E}_x[{\rm log}~p_t], \ t = 1, 2, ..., L.
\end{equation}

Although OIM effectively employs both labeled and unlabeled samples, we still observe two limitations. First, the distances are only computed between the input features and the features stored in the lookup table and circular queue, while no comparisons are made between the input features. Second, the log-likelihood loss term does not give an explicit distance metric between feature pairs. 

\begin{figure}[t]
\centering
\includegraphics[width=\linewidth]{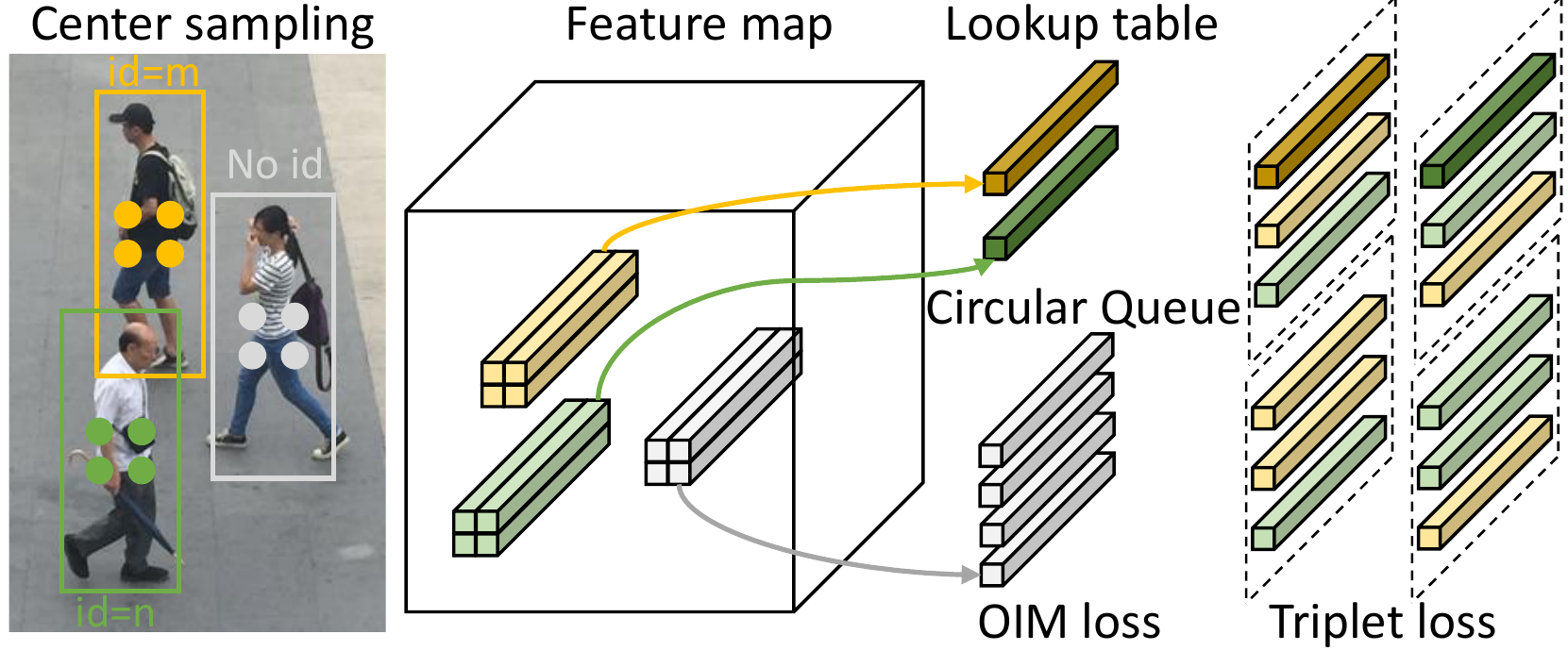}
\caption{Illustration of the Triplet-aided Online Instance Matching loss, where both the features from the input image and the lookup table are sampled to form the triplet.}
\label{fig:loss}
\end{figure}

To improve OIM, we propose a specifically designed triplet loss. For each person in the input images, we employ the center sampling strategy as in~\cite{DBLP:journals/tip/KongSLJLS20}. As shown in Fig.~\ref{fig:loss}, for each person, a set of features located around the person center are considered as positive samples. The objective is to pull the feature vectors from the same person close, and push the vectors from different people away. Meanwhile, the features from the labeled persons should be close to the corresponding features stored in the LUT, and away from the other features in the LUT.

More specifically, suppose we sample $S$ vectors from one person; we get $X_m = \{x_{m,1},..., x_{m,S}, v_m\}$ and $X_n=\{x_{n,1},..., x_{m,S}, v_n\}$ as the candidate feature sets for the persons with identity labels $m$ and $n$, respectively, where $x_{i,j}$ denotes the $j$-th feature of person $i$, and $v_i$ is the $i$-th feature in the LUT. Given $X_m$ and $X_n$, positive pairs can be sampled within each set, while negative pairs are sampled between the two sets. The triplet loss can be calculated as:
\begin{equation}
    \mathcal{L}_{\text{tri}} = \sum_{\text{pos,~neg}}[M + D_{\text{pos}} - D_{\text{neg}}],
\end{equation}
where $M$ denotes the distance margin, and $D_{\text{pos}}$ and $D_{\text{neg}}$ denote the Euclidean distances between the positive pair and the negative pair, respectively. Finally, the Triplet-aided OIM (TOIM) loss is the summation of these two terms:
\begin{equation}
    \mathcal{L}_{\text{TOIM}} = \mathcal{L}_{\text{tri}}+\mathcal{L}_{\text{OIM}}.
\end{equation}

\begin{figure*}[t]
\centering
\subfloat[Training pipeline of ROI-AlignPS\label{subfig-r-1}]{%
   \includegraphics[width=0.485\linewidth]{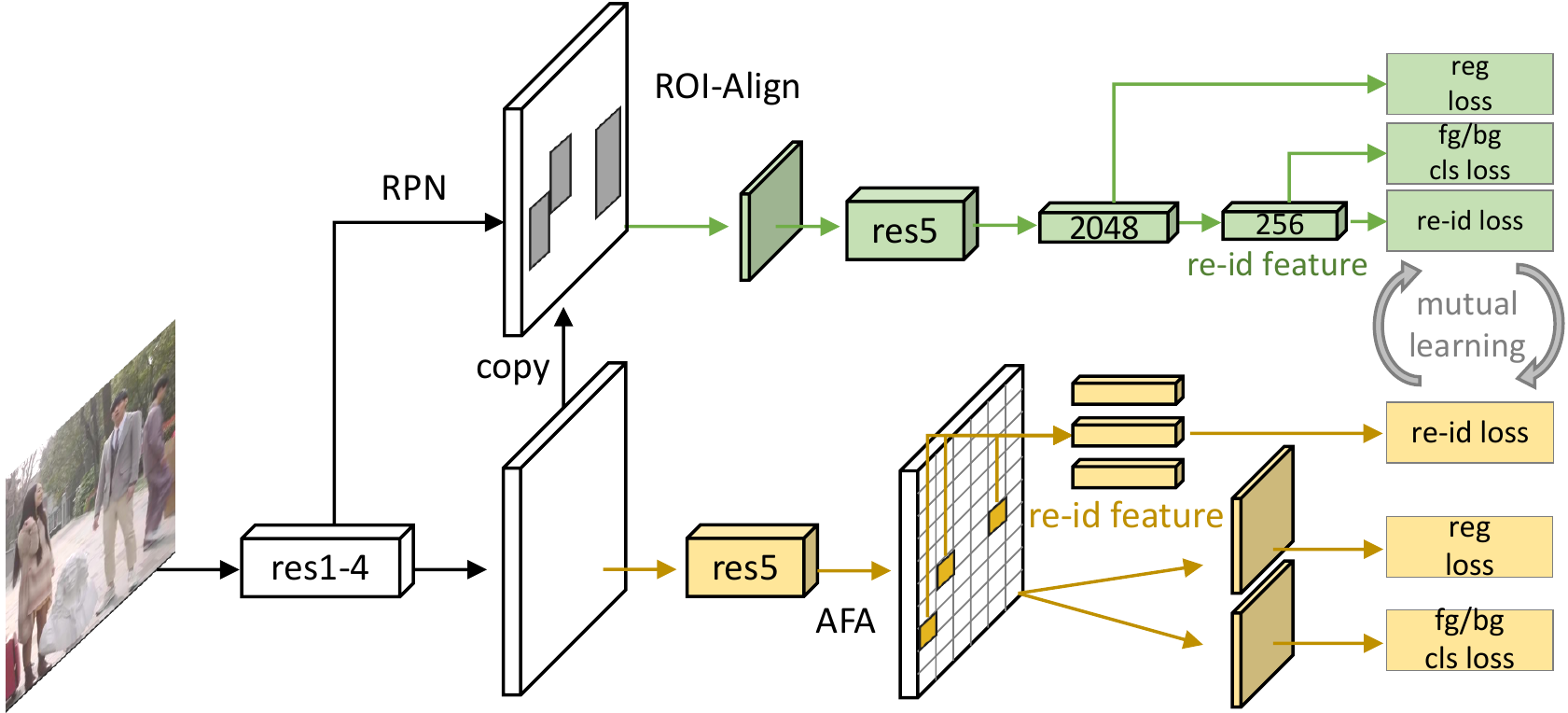}
}
\hspace{1mm}
\centering
\subfloat[Test pipeline of ROI-AlignPS\label{subfig-r-2}]{%
   \includegraphics[width=0.485\linewidth]{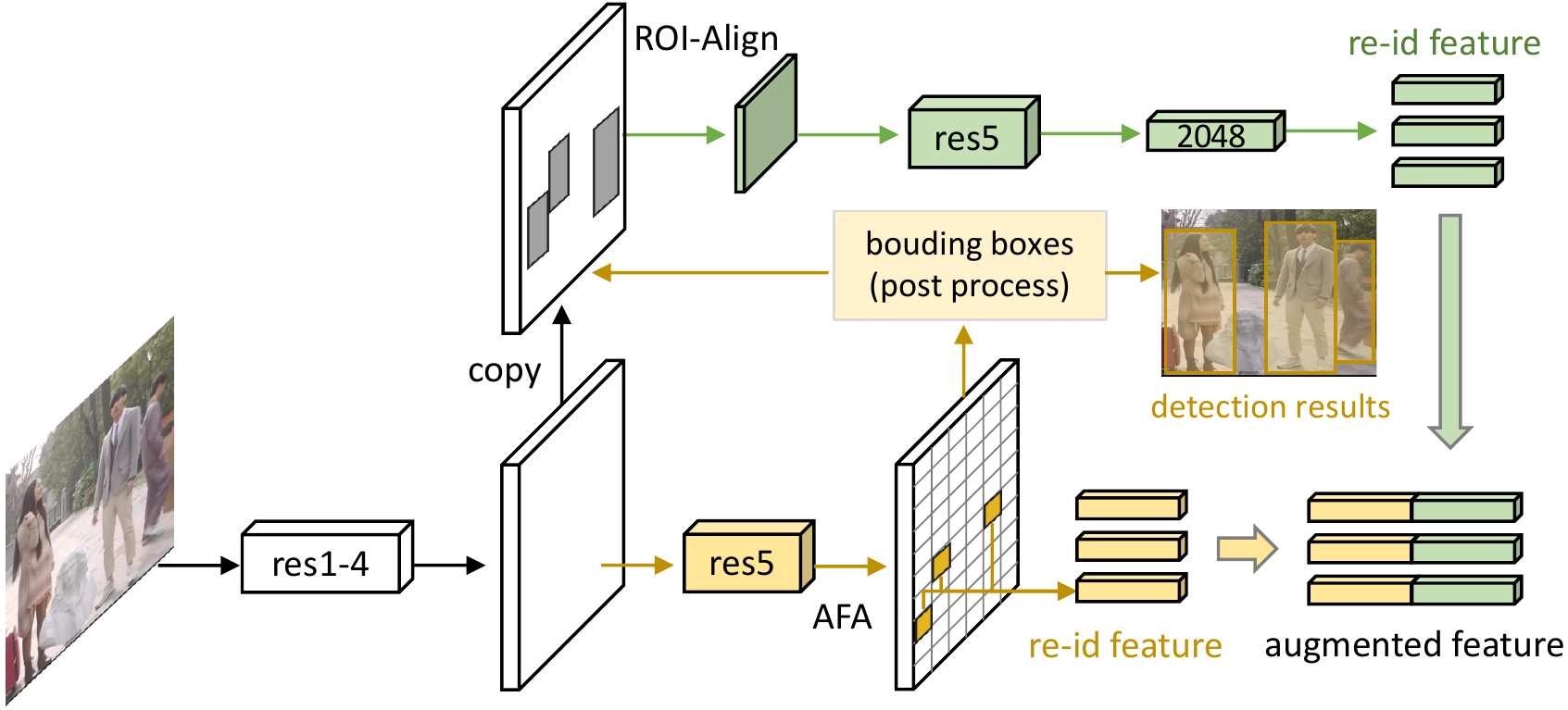}
}
 \caption{The architecture of ROI-AlignPS, which can be regarded as AlignPS augmented with a Faster R-CNN branch.}
 \label{fig:ralignps}
\end{figure*}

\subsection{AlignPS Augmented with Explicit Region Alignment}\label{sec:mls}
\textbf{ROI-AlignPS}. Although the proposed AFA module implicitly addresses the region misalignment issue with deformable convolution, it is still desirable to investigate the explicitly aligned learning schemes, e.g., cropping pedestrians for re-id learning~\cite{DBLP:conf/cvpr/ZhengZSCYT17,DBLP:conf/eccv/ChenZOYT18} and the ROI-Align operation in Faster R-CNN~\cite{DBLP:conf/cvpr/XiaoLWLW17,DBLP:conf/cvpr/ChenZYS20}. However, learning based on the cropped pedestrians inevitably transfers the whole framework into a two-step model, which significantly downgrades the efficiency. In the meantime, the dense region proposals in Faster R-CNN also make it inefficient during inference. To address these issues, we propose a novel variant, which we name ROI-AlignPS, by augmenting our efficient anchor-free framework with an explicit region alignment module. 

In the training phase, two parallel branches are simultaneously trained with implicit and explicit region alignment, respectively. As shown in Fig.~\ref{subfig-r-1}, the bottom branch is the original AlignPS model, while the top branch follows the architecture of NAE~\cite{DBLP:conf/cvpr/ChenZYS20}, which includes a RPN and ROI-Align operation to output the re-id features of a certain region. Both branches share the ResNet-50 backbone from \textit{res1} to \textit{res4} layers, while having separate \textit{res5} layers. In the test phase, as illustrated in Fig~\ref{subfig-r-2}, the bottom AlignPS branch first outputs bounding boxes and re-id features. Then, the processed bounding boxes are fed to the top branch to extract explicitly aligned re-id features corresponding to each person. Finally, the re-id features from these two branches are combined to generate a more robust representation.

Although ROI-AlignPS contains an anchor-free branch and an anchor-based branch, the dense anchors in Faster R-CNN are only involved in the training stage. During inference, RPN is discarded and only a small number of detected bounding boxes are processed with ROI-Align. In this way, ROI-AlignPS can still remain highly efficient during the inference stage. Moreover, the explicit and implicit region alignment strategies are complementary to each other. Specifically, the ROI-Align branch extracts re-id features based on bounding boxes, while AlignPS dynamically adapts to certain regions. Meanwhile, the performance of both methods may be influenced by occlusions and backgrounds. Therefore, combining these two kinds of features intuitively reduces such impacts, and the effectiveness of this design is validated in Sec~\ref{sec:ana-r}.

\textbf{Branch-Level Mutual Learning}. Prior works~\cite{DBLP:journals/pami/LiH18a,DBLP:conf/iccv/PengLZLQT19,Hong_2021_CVPR,Dai_2021_CVPR} found that feature interaction and knowledge distillation can benefit feature learning in neural networks. In our case, it would be desirable if the prediction of the two branches can reach a better consensus. To fulfill this, we investigate several branch interaction strategies as follows.

\textit{(1) Mutual Information Maximization}. Mutual information is defined to measure the dependency between two random variables $X$ and $Y$:
\begin{equation}
    \mathcal{I}(X,Y) = \int_{X}\int_{Y}p(x,y) {\rm log}\frac{p(x,y)}{p(x)p(y)}dxdy,
\end{equation}
where $p(x,y)$ is the joint probability density function, $p(x)$ and $p(y)$ are the marginal probability density functions of $X$ and $Y$, respectively. As demonstrated in prior works~\cite{Kinney3354,DBLP:conf/iclr/HjelmFLGBTB19}, mutual information is able to depict the mutual dependence between $X$ and $Y$, no matter how nonlinear the dependence is. Thus, mutual information is more flexible than the measurement of correlation. In this work, random variables $X$ and $Y$ represent the re-id features from the two branches in ROI-AlignPS, respectively. We aim to maximize the mutual information between $X$ and $Y$, such that the overall representation will focus on the identity information. As it is non-trivial to directly calculate the mutual information, we employ a neural estimator~\cite{DBLP:conf/icml/BelghaziBROBHC18} to maximize the lower bound of the mutual information. Specifically, 
\begin{equation}
    \mathcal{I}_{\Theta}(X,Y)  =  \sup_{\theta\in\Theta} \mathbb{E}_{\mathbb{P}_{{X}{Y}}}[T_\theta] - \log(\mathbb{E}_{\mathbb{P}_{{X}\otimes{Y}}}[e^{T_\theta}]),
\end{equation}
where $\mathbb{P}_{{X}{Y}}$ is the joint distribution and $\mathbb{P}_{{X}\otimes{Y}}$ represents the marginal distribution, $T_{\theta}$ is a neural network parameterized by $\theta \in \Theta$. To maximize $\mathcal{I}_{\Theta}(X,Y) $, we minimize its negative value:
\begin{equation}
    \mathcal{L}_{mi} = -\mathcal{I}_{\Theta}(X,Y).
\end{equation}

\textit{(2) KL-Divergence Minimization}. 
The Kullback-Leibler (KL) divergence is widely employed to measure the difference between two distributions, including successful applications in feature distillation in person search~\cite{DBLP:conf/cvpr/DongZST20a,DBLP:conf/aaai/ZhangWBSY21}. In our case, we minimize the KL-divergence between identity predictions of the two branches. Suppose $p^{A}$ and $p^R$ denote the output probabilities of the AlignPS branch and the ROI-Align branch, respectively, the KL-divergence is calculated as:
\begin{align}
    \mathcal{L}_{kl} = {\rm KL} (p^A \| p^R) 
                     = \sum_{i=1}^{L} p_i^A {\rm log} \frac{p_i^A}{p_i^R},
\end{align}
where $p_i^A$ and $p_i^R$ are calculated from Eq.~\ref{eq:one}, and $L$ is the number of identities in the training set. In this way, the output features could reach better prediction-level consensus.

\textit{(3) Diversity Maximization}. Rather than pursuing a consensus between the two branches, we aim to diversify the features belonging to the same identity, to yield more robust ensemble results.
To enhance diversity, we promote the output features of the two branches to be different. Specifically, suppose $x_i^A$ and $x_i^R$ denote the features of the $i$-th person from the AlingPS branch and ROI-Align branch, respectively, we aim to minimize the cosine similarity between the corresponding features:
\begin{equation}
    \mathcal{L}_{dv} = \frac{1}{B}\sum_{i=1}^{B} {\rm cos}(x_i^A, x_i^R),
\end{equation}
where $B$ denotes the number of persons in a mimi-batch, and $\rm cos$ denotes the cosine similarity.

\begin{figure*}[t]
\centering
\subfloat[Comparative results on CUHK-SYSU\label{subfig-b-1}]{%
   \includegraphics[width=0.48\linewidth]{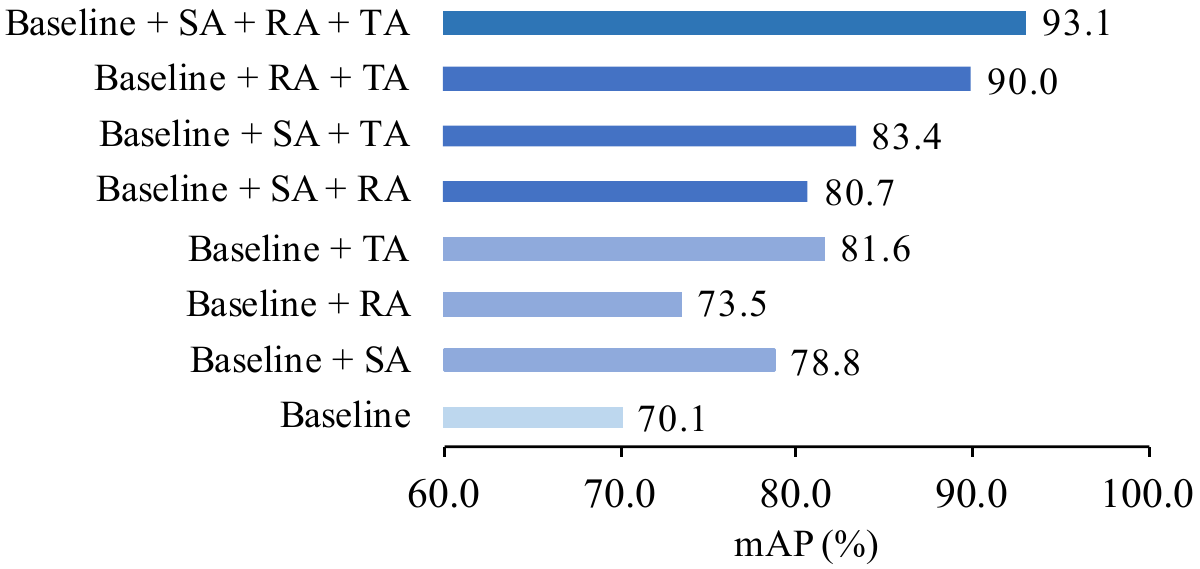}
}
\hspace{2mm}
\centering
\subfloat[Comparative results on PRW\label{subfig-b-2}]{%
   \includegraphics[width=0.48\linewidth]{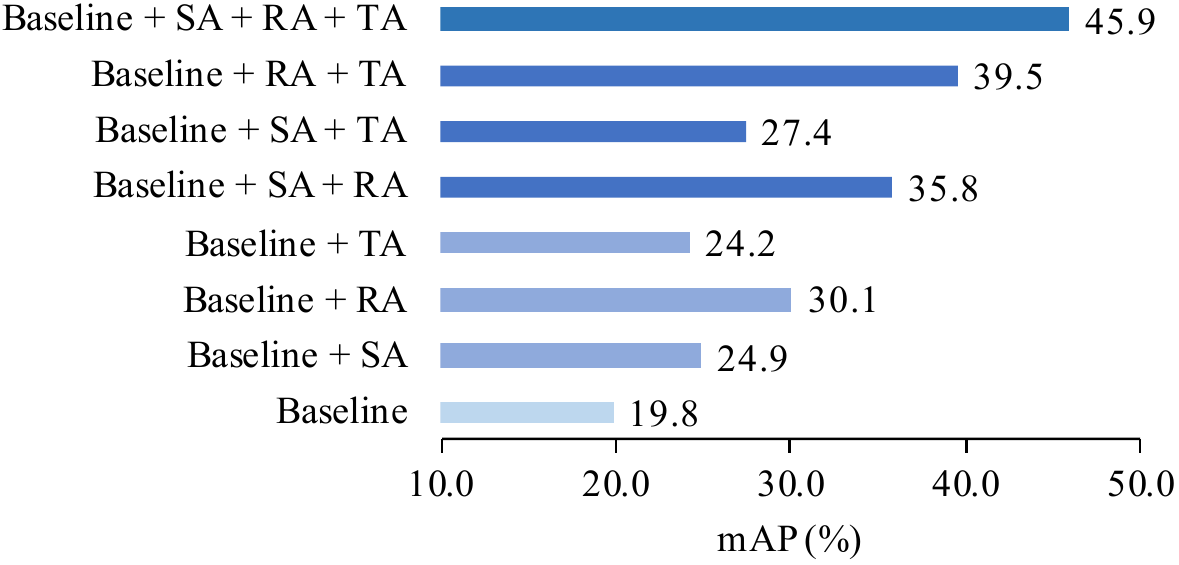}
}
\caption{Comparative results on CUHK-SYSU and PRW with different alignment strategies, i.e., scale alignment (SA), region alignment (RA), and task alignment (TA).}
\label{fig:baseline}
\end{figure*}

\textbf{Discussions}. A recent work, SeqNet~\cite{DBLP:conf/aaai/LiM21}, proposes a sequential architecture based on Faster R-CNN, which first predicts the detection results, and then extracts re-id features with an additional ROI-Align branch. Although ROI-AlignPS shares certain spirits with SeqNet, there exist several significant differences. (1) The motivation of SeqNet is to improve the quality of proposals, such that the re-id features could be better learned. In ROI-AlignPS, rather than considering the detection issue, the ROI-Align branch serves as an explicitly aligned re-id feature extractor, to complement the features from the AlignPS branch. (2) Despite the strong performance, SeqNet relies on a post graph matching technique. In contrast, the re-id features in ROI-AlignPS could be mutually promoted and the ensemble results are highly discriminative without post-processing. (3) In the inference stage, ROI-AlignPS still runs in an anchor-free way. Consequently, our framework is more efficient than SeqNet. The comparative results between ROI-AlignPS and SeqNet can be found in Sec.~\ref{sec:cmps}. 

\section{Experiments}

\subsection{Datasets and Settings}
\textbf{CUHK-SYSU}~\cite{DBLP:conf/cvpr/XiaoLWLW17} is a large-scale person search dataset which contains 18,184 images, with 8,432 different identities and 96,143 annotated bounding boxes. The images come from two kinds of data sources (i.e., real street snaps and movies/TV),
covering diverse scenes and including variations of viewpoints, lighting, resolutions, and occlusions.
We utilize the standard training/test split, where the training set contains 5,532 identities and 11,206 images, and the test set contains 2,900 query persons and 6,978 images. This dataset also defines a set of protocols with gallery sizes ranging from 50 to 4,000. We report the results using the default gallery size of 100 unless otherwise specified.

\textbf{PRW}~\cite{DBLP:conf/cvpr/ZhengZSCYT17} was captured using six static cameras in a university campus.
All the images are extracted from the surveillance videos, which consist of 11,816 video frames in total. Person identities and bounding boxes are manually annotated, resulting in 932 labeled persons with 43,110 bounding boxes. The dataset is split into a training set of 5,704 images with 482 different identities, and a test set of 2,057 query persons and 6,112 images. Results are reported based on this split.

\textbf{Evaluation Metric}. We employ the mean average precision (mAP) and top-1 accuracy to evaluate the performance for person search. We also employ recall and average precision (AP) to measure the detection performance.

\subsection{Implementation Details}
We employ ResNet-50~\cite{DBLP:conf/cvpr/HeZRS16} pretrained on ImageNet~\cite{DBLP:conf/cvpr/DengDSLL009} as the backbone. 
We set the batch size to 4, and adopt the stochastic gradient descent (SGD) optimizer with weight decay of 0.0005. The initial learning rate is set to 0.001 and is reduced by a factor of 10 at epoch 16 and 22, with a total of 24 epochs. We use a warmup strategy for 300 steps. We employ a multi-scale training strategy, where the longer side of the image is randomly resized from 667 to 2000 during training, while zero padding is utilized to fit the images with different resolutions. For inference, we rescale the test images to a fixed size of 1500$\times$900. Following~\cite{DBLP:conf/aaai/ChenZO0S20}, we add a focal loss~\cite{DBLP:conf/iccv/LinGGHD17} to the original OIM loss.
All the experiments are implemented based on PyTorch~\cite{DBLP:conf/nips/PaszkeGMLBCKLGA19} and MMDetection~\cite{DBLP:journals/corr/abs-1906-07155}, with an NVIDIA Tesla V100 GPU. It takes around 38 and 24 hours for ROI-AlignPS to finish training on CUHK-SYSU and PRW, respectively.

\subsection{Analysis of AlignPS}\label{sec:analytical}

\textbf{Baseline}.
We directly add a re-id head in parallel with the detection head to the FCOS model and take it as our baseline. As shown in Fig.~\ref{fig:baseline}, each of the alignment strategies brings notable improvements to the baseline, and combining all of them yields 23\% and 26.1\% improvements in mAP on CUHK-SYSU and PRW, respectively.

\begin{figure*}[t]
\begin{center}
\subfloat[Deformable conv at lateral $C_3$ layer in AFA\label{subfig-dcnvis-1}]{%
   \includegraphics[width=\linewidth]{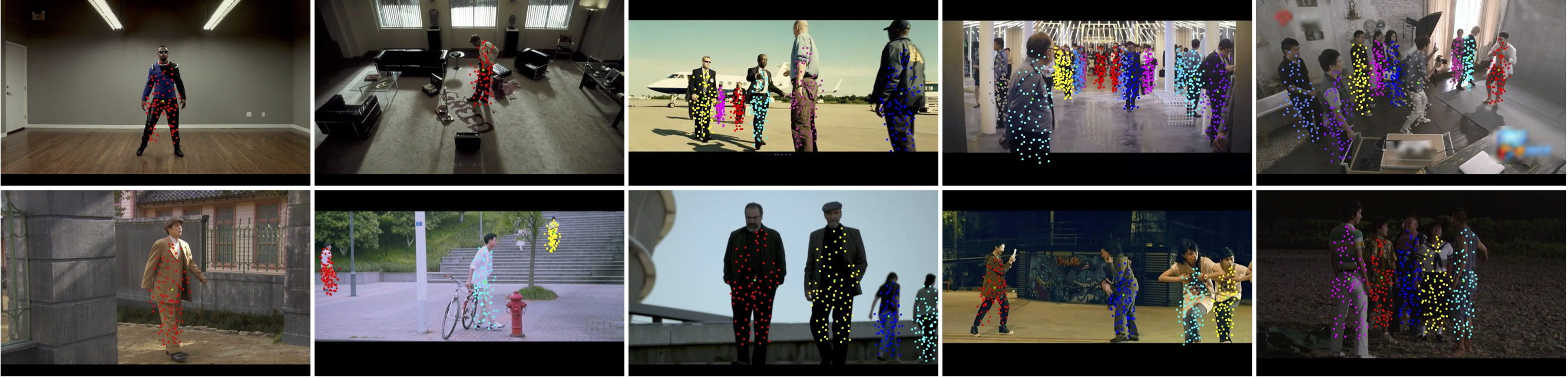}
}
 \hfill
\subfloat[Deformable conv at lateral $C_4$ layer in AFA\label{subfig-dcnvis-2}]{%
   \includegraphics[width=\linewidth]{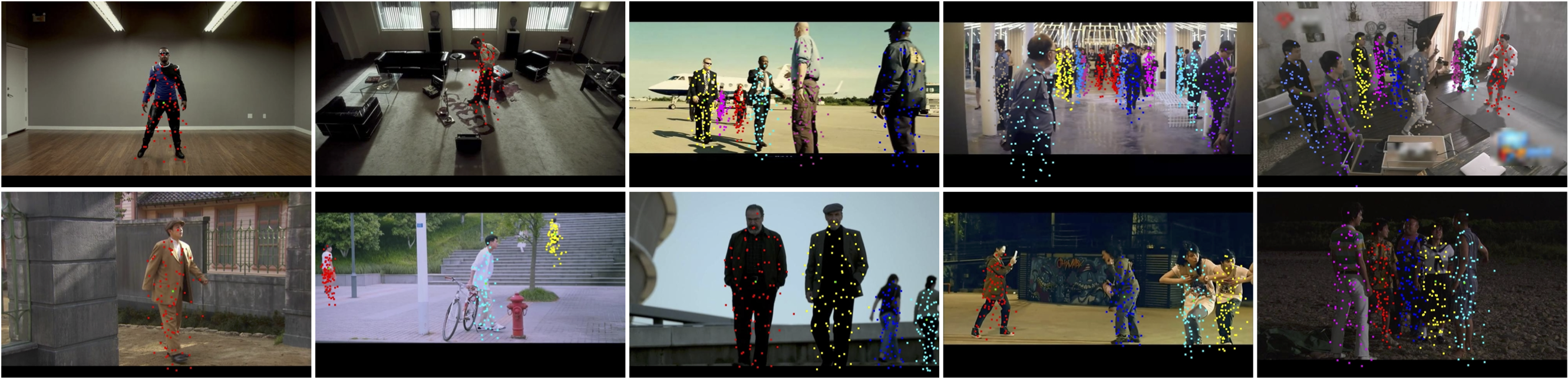}
}
\end{center}
   \caption{Each image shows the sampling locations of two levels of 3$\times$3 ($9^2 = 81$ points at each location) deformable filters: (a) Lateral deformable conv $C_3$ + Output deformable conv; (b) Lateral deformable conv $C_4$ + Output deformable conv. We illustrate different locations with different colors, while center locations of people are marked in green. Please zoom in for better visualization.}
\label{fig:dcnvis}
\end{figure*}

\begin{table}[ht]
\small
\centering
\begin{tabular}{p{2cm}|p{1.1cm}<{\centering}p{1.1cm}<{\centering}|p{1.1cm}<{\centering}p{1.1cm}<{\centering}}
\hline\thickhline
\rowcolor{mygray} 
  & \multicolumn{2}{c|}{Detection} & \multicolumn{2}{c}{Re-id}   \\ \cline{2-5} 
\rowcolor{mygray} 
\multirow{-2}{*}{CUKH-SYSU}  & Recall & AP  & mAP  & top-1  \\  \hline \hline     
$P_3$
  & 90.3    & \textbf{81.2} &\textbf{93.1} & \textbf{93.4}  \\
$P_4$  & 87.5   & 78.7       & 92.7   & 93.1    \\ 
$P_5$  & 79.0   & 71.7       & 89.3   & 89.5  \\
$P_3$, $P_4$  & 90.4  & 80.5       & 91.1 & 91.6 \\ 
$P_3$, $P_4$, $P_5$  & \textbf{90.9}  & 80.4       & 90.0 & 90.5 \\\hline \hline
\rowcolor{mygray} 
  & \multicolumn{2}{c|}{Detection} & \multicolumn{2}{c}{Re-id}   \\ \cline{2-5} 
\rowcolor{mygray} 
\multirow{-2}{*}{PRW}  & Recall & AP  & mAP  & top-1  \\  \hline \hline     
$P_3$
  &  94.8   & 92.9  &\textbf{45.9} & \textbf{81.9}  \\
$P_4$  &   93.4  &  91.8      & 41.4   & 77.5    \\ 
$P_5$  &  85.7  &    83.7    & 25.3  & 57.4  \\
$P_3$, $P_4$  & 94.7 & 92.9      & 40.8 & 77.1 \\ 
$P_3$, $P_4$, $P_5$  &  \textbf{95.4}  &    \textbf{93.5}  & 39.5 & 74.3 \\\hline
\end{tabular}
\caption{Comparative results on CUHK-SYSU and PRW by employing different levels of features. $P_3$, $P_4$, and $P_5$ are the feature maps with strides of 8, 16, and 32, respectively. }
\label{tab:scale}
\end{table}

\textbf{Scale Alignment}. 
To evaluate the effects of scale alignment, we employ feature maps from different levels of AFA and report the results in Table~\ref{tab:scale}. Specifically, we evaluate the features from $P_3$, $P_4$, and $P_5$ with strides of 8, 16, and 32, respectively. As can be observed, features from the largest scale $P_3$ yield the best performance on both CUHK-SYSU and PRW, due to the fact that they absorb different levels of features from AFA, providing richer information for detection and re-id. Similar to FCOS, we also evaluate the performance by assigning people of different scales to different feature levels. We set the size ranges for \{$P_3, P_4\}$ as [0, 128] and [128, $\infty$], while the prediction ranges for \{$P_3, P_4, P_5\}$ are [0, 128], [128, 256], and [256, $\infty$], respectively. We can see that these dividing strategies achieve slightly better detection results w.r.t. the recall rate. However, they bring back the scale misalignment issue to person re-id, resulting in worse performance compared with single-scale features. Also note that this issue is not well addressed with the multi-scale training strategy. All the above results demonstrate the necessity and effectiveness of the proposed scale alignment strategy.


\begin{table}[t]
\small
\centering
\begin{tabular}{p{0.8cm}<{\centering}p{0.8cm}<{\centering}p{0.8cm}<{\centering}|p{0.8cm}<{\centering}p{0.8cm}<{\centering}|p{0.8cm}<{\centering}p{0.8cm}<{\centering}}
\hline\thickhline
\rowcolor{mygray}  
Lateral              & Output               & Feature              & \multicolumn{2}{c}{CUHK-SYSU}     & \multicolumn{2}{|c}{PRW}                   \\ \cline{4-7} 
\rowcolor{mygray}  
dconv                 & dconv                 & concat               & mAP                  & \multicolumn{1}{c}{top-1} & \multicolumn{1}{|c}{mAP}                  & \multicolumn{1}{c}{top-1} \\ 
\hline \hline  
 &   & & 83.4 & 83.7              & 27.4       & 62.3            \\ 
$\checkmark$ &   & & 90.6 & 90.8        &39.4   &   74.7             \\
 & $\checkmark$  & & 91.4 & 91.9          & 40.6   &    77.0              \\
  &   & $\checkmark$ & 84.0 & 84.1         & 29.1  &     62.4              \\
$\checkmark$ & $\checkmark$  & & 91.8 & 92.2     & 42.8  &   80.1                    \\
$\checkmark$ &   & $\checkmark$ & 90.7 & 91.0     & 39.6  &   75.3                     \\
 & $\checkmark$  & $\checkmark$& 92.0 & 92.5       & 40.7  &    77.5                  \\
$\checkmark$ & $\checkmark$  & $\checkmark$ & \textbf{93.1} & \textbf{93.4}    &  \textbf{45.9} &    \textbf{81.9}                  \\\hline
\end{tabular}
\caption{Comparative results on CUHK-SYSU and PRW by employing different components in AFA for region alignment. ``dconv'' stands for deformable convolution. }
\label{tab:region}
\end{table}

\textbf{Region Alignment}. 
We conduct experiments with different combinations of lateral deformable conv, output deformable conv and feature concatenation, and analyze how different region alignment components influence the overall performance. The results are reported in Table~\ref{tab:region}. Without all these modules, the framework only achieves 83.4\% and 27.4\% in mAP on CUHK-SYSU, and PRW, respectively, which is 9.7\% and 18.5\% lower than the full model. The individual components of lateral deformable conv and output deformable conv improve the model by $\sim$7\% and $\sim$8\% on CUHK-SYSU, respectively. Feature concatenation also brings $\sim$1\% improvements. By combining two of the three components, we observe consistent improvements. Finally, employing all the three modules yields 93.1\% in mAP and 93.4\% in top-1 accuracy on CUHK-SYSU, significantly boosting the performance. On PRW, we also observe consistent improvement by introducing these strategies. These ablation studies thoroughly demonstrate the effectiveness of region alignment.

To further illustrate how the deformable convolutions work in our framework, we visualize the learned offsets of the deformable filters in Fig.~\ref{fig:dcnvis}. We observe that the proposed framework is capable of learning adaptive receptive field according to the layout of the human body, and is robust to occlusion, crowding, and scale/illumination variations. We also observe that the lateral deformable conv in $C_3$ learns tighter offsets around the body center, while the offsets in the $C_4$ layer cover larger regions, which makes the two layers complementary to each other. 

\begin{table}[t]
\small
\centering
\begin{tabular}{p{1.9cm}|p{1.1cm}<{\centering}p{1.1cm}<{\centering}|p{1.1cm}<{\centering}p{1.1cm}<{\centering}}
\hline\thickhline
\rowcolor{mygray} 
  & \multicolumn{2}{c|}{Detection} & \multicolumn{2}{c}{Re-id}   \\ \cline{2-5} 
\rowcolor{mygray} 
\multirow{-2}{*}{CUHK-SYSU}  & Recall & AP  & mAP  & top-1  \\  \hline \hline     
$T_1$
   & 87.5   & 79.0       & 80.3   & 79.2   \\
$T_2$  & 89.1   & 78.6      & 77.1   & 75.9    \\ 
$T_3$  & 90.1   & 81.4       & 80.7   & 80.2  \\
AlignPS  & 90.3    & 81.2 &\textbf{93.1} & \textbf{93.4}\\\hline \hline
\rowcolor{mygray} 
  & \multicolumn{2}{c|}{Detection} & \multicolumn{2}{c}{Re-id}   \\ \cline{2-5} 
\rowcolor{mygray} 
\multirow{-2}{*}{PRW}  & Recall & AP  & mAP  & top-1  \\  \hline \hline        
$T_1$
   &  94.3   &  92.5       & 35.5   & 67.4   \\
$T_2$  & 96.9     & 92.8     & 31.8   & 64.3    \\ 
$T_3$  &  94.5  &  92.6     & 35.8   & 69.1\\
AlignPS  &  94.8   & 92.9  &\textbf{45.9} & \textbf{81.9}\\\hline
\end{tabular}
\caption{Comparative results on CUHK-SYSU and PRW with different training structures. }
\label{tab:tasks}
\end{table}

\begin{figure}[t]
\vspace{-5mm}
\centering
\subfloat[$T_1$\label{subfig-2-1}]{%
   \includegraphics[width=0.485\linewidth]{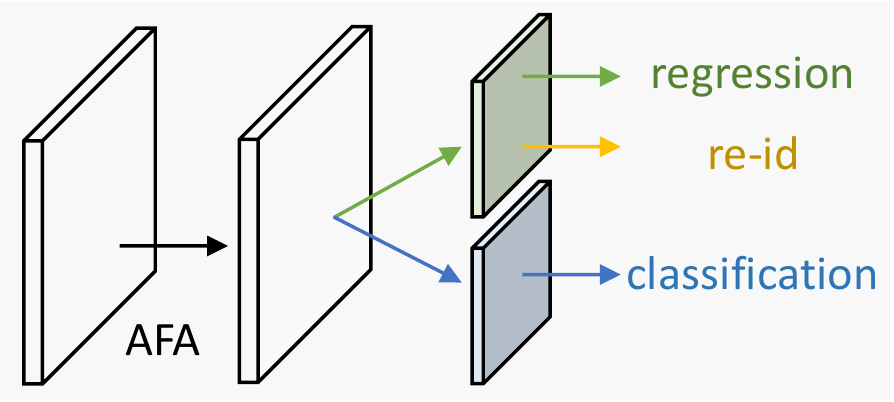}
}
 \hfill
\subfloat[$T_2$\label{subfig-2-2}]{%
   \includegraphics[width=0.485\linewidth]{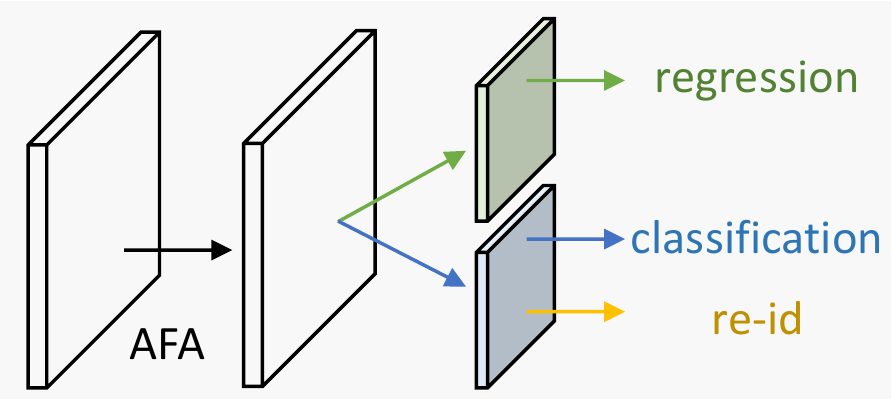}
}
 \hfill
\subfloat[$T_3$\label{subfig-2-3}]{%
   \includegraphics[width=0.485\linewidth]{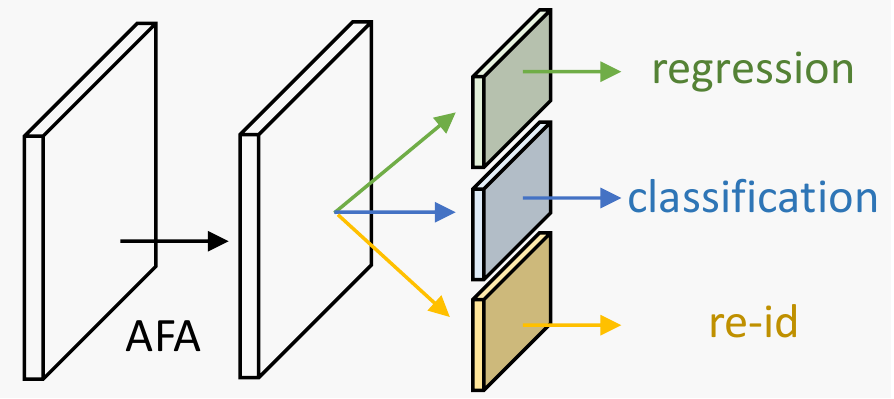}
}
 \hfill
\subfloat[AlignPS\label{subfig-2-4}]{%
   \includegraphics[width=0.485\linewidth]{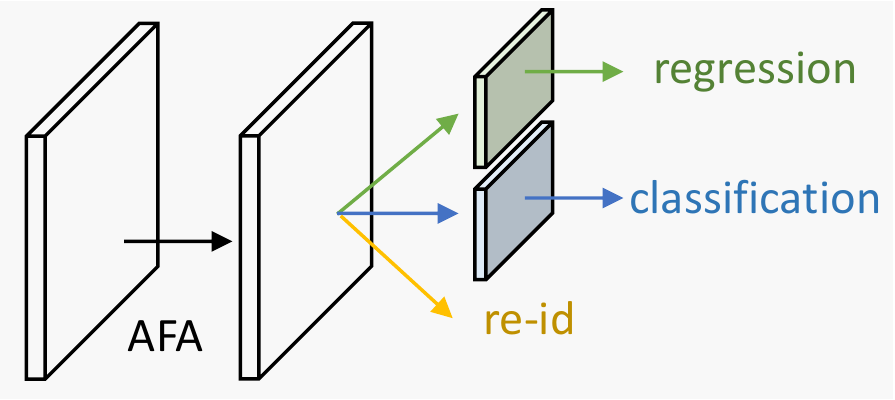}
}
 \caption{Illustration of different structures for training the detection and re-id tasks.}
 \label{fig:tasks}
\end{figure}

\textbf{Task Alignment}.
Since person search aims to simultaneously address detection and re-id subtasks in a single framework, it is important to understand how different configurations of the two subtasks influence the overall task and which subtask should be paid more attention to. To this end, we design several structures to compare different training options (as shown in Fig.~\ref{fig:tasks}), the performance of which is summarized in Table~\ref{tab:tasks}. As can be observed, the structures of $T_1$ and $T_2$, where re-id features are shared with the regression and classification heads, respectively, yield significantly lower performance in re-id compared with our design. This indicates that the detection task takes advantage of the shared heads. As for $T_3$ where re-id and detection have independent feature heads, it achieves slightly better performance compared with $T_1$ and $T_2$, but still remarkably underperforms our design. 
These results indicate that our ``re-id first'' structure achieves the best task alignment among all these designs.

\begin{table}[t]
\small
\centering
\begin{tabular}{p{2.4cm}|p{0.8cm}<{\centering}p{0.8cm}<{\centering}|p{1.1cm}<{\centering}p{1.2cm}<{\centering}}
\hline\thickhline
\rowcolor{mygray} 
{CUHK-SYSU}  & mAP & top-1  & $\Delta$ mAP  & $\Delta$ top-1  \\  \hline \hline     
OIM
   & 92.4   & 92.9       & -   & -  \\
TOIM w/o LUT  & 92.8   & 93.2       & +0.4   & +0.3   \\ 
TOIM  w/ LUT  &\textbf{93.1} & \textbf{93.4} & +0.7    & +0.5\\\hline
\rowcolor{mygray} 
{PRW}  & mAP & top-1  & $\Delta$ mAP  & $\Delta$ top-1  \\  \hline \hline     
OIM
   & 45.7   & 81.8       & -   & -  \\
TOIM w/o LUT  & 45.8   & 81.8       & +0.1   & +0   \\ 
TOIM  w/ LUT  &\textbf{45.9} & \textbf{81.9} & +0.2    & +0.1\\\hline
\end{tabular}
\caption{Comparative results on CUHK-SYSU and PRW with different loss functions. }
\label{tab:toim}
\end{table}

\begin{table}[t]
\small
\centering
\begin{tabular}{p{1.9cm}|p{2.7cm}|p{1.1cm}<{\centering}p{1.1cm}<{\centering}}
\hline\thickhline
\rowcolor{mygray} 
{CUHK-SYSU}  & Deformable conv   &  mAP  &  top-1  \\  \hline \hline     
ResNet-50
   & none       & 93.1   & 93.4  \\
ResNet-50
   & res3      & 93.5   & 93.9  \\
ResNet-50
   & res3{\,}\&{\,}res4      & 93.5   & 94.0  \\
ResNet-50
   & res3{\,}\&{\,}res4{\,}\&{\,}res5     &\textbf{94.0} & \textbf{94.5} \\
\hline \hline
\rowcolor{mygray} 
{PRW}  & Deformable conv   &  mAP  &  top-1  \\  \hline \hline     
ResNet-50
   & none       & 45.9   & 81.9  \\
ResNet-50
   & res3      & 45.8  & 81.9  \\
ResNet-50
   & res3{\,}\&{\,}res4      & 45.9   & 82.1  \\
ResNet-50
   & res3{\,}\&{\,}res4{\,}\&{\,}res5     &\textbf{46.1} & \textbf{82.1} \\
\hline
\end{tabular}
\caption{Comparative results on CUHK-SYSU and PRW with different deformable conv layers in the backbone model. }
\label{tab:dcn}
\end{table}

\textbf{TOIM Loss}.
We evaluate the performance of our framework when adopting different loss functions and report the results in Table~\ref{tab:toim}. We find that directly employing a triplet loss brings slight improvement. When employing the items in the LUT, the TOIM improves the mAP and top-1 accuracy on CUHK-SYSU by 0.7\% and 0.5\%, respectively. These two terms also slightly improve the performance on PRW. This indicates that it is beneficial to consider the relations between the input features and the features stored in the LUT.


\textbf{Deformable Conv in the Backbone.}
As shown in Table~\ref{tab:dcn}, inserting deformable convolutions into the backbone network has positive effects on our framework. However, the contribution of the deformable conv layers in the backbone network is less significant than the deformable conv layers in our AFA module, e.g., with all the res3{\,}\&{\,}res4{\,}\&{\,}res5 deformable conv layers, only $\sim$1\% and 0.2\% improvements are observed on CUHK-SYSU and PRW, respectively. These results indicate that the proposed AFA works as the key module for successful feature alignment.



\subsection{Analysis of ROI-AlignPS}\label{sec:ana-r}
We analyze the effectiveness of ROI-AlignPS by answering the following questions.

\begin{table}[t]
\small
\centering
\begin{tabular}{p{2.2cm}|p{0.8cm}<{\centering}p{0.8cm}<{\centering}|p{0.8cm}<{\centering}p{0.8cm}<{\centering}|p{0.7cm}<{\centering}}
\hline\thickhline
\rowcolor{mygray} \rowcolor{mygray} 
 {Separate} & \multicolumn{2}{c|}{CUHK-SYSU} & \multicolumn{2}{c|}{PRW}  & \multicolumn{1}{c}{Time}   \\ \cline{2-5} 
\rowcolor{mygray} 
{Training}  & mAP & top-1  & mAP  & top-1 & (ms) \\  \hline \hline   
ROI-Align
   & 92.0   & 92.3       & 43.1   & 80.7 & 83\\
AlignPS  & 93.1   & 93.4       & 45.9   & 81.9  & 61\\ 
Concatenation & 93.0 & 93.6 & 44.5    & 81.3 & 144\\\hline
\rowcolor{mygray} \rowcolor{mygray} 
 {Joint Training}  & \multicolumn{2}{c|}{CUHK-SYSU} & \multicolumn{2}{c|}{PRW}  &{Time}  \\ \cline{2-5} 
\rowcolor{mygray} 
{w/o RPN}  & mAP & top-1  & mAP  & top-1 & (ms) \\  \hline \hline    
ROI-Align
   & 90.4   &   91.3     & 43.9   & 81.5 & 83\\
AlignPS  & 93.2   & 92.8       &44.9   & 80.0  & 61\\ 
Concatenation  &94.3 & 94.8 & 48.6    & 83.2 & 75\\ \hdashline
ROI-Align$^{*}$
   & 93.4   & 93.8       & 47.5   & 84.5 & 75\\
\hline
\rowcolor{mygray} \rowcolor{mygray} 
{Joint Training}  & \multicolumn{2}{c|}{CUHK-SYSU} & \multicolumn{2}{c|}{PRW}  &{Time}  \\ \cline{2-5} 
\rowcolor{mygray} 
{w/ RPN}  & mAP & top-1  & mAP  & top-1 & (ms) \\  \hline \hline     
ROI-Align
   & 93.3   &   93.9     & 47.8   & 83.3 & 83\\
AlignPS  & 94.0   & 94.1       &46.4   & 81.2  & 61\\ 
Concatenation  &\textbf{95.0} & \textbf{95.3} & \textbf{50.3}    & \textbf{84.3} & 75\\\hline
\end{tabular}
\caption{Comparative results with different structures. ROI-Align$^{*}$ denotes the model trained without re-id feature in the AlignPS branch, i.e., AlignPS concentrates on the detection task while ROI-Align focuses on re-id. This also yields an efficient anchor-free framework with fair performance. }
\label{tab:rbn}
\end{table}

\textbf{Is the ROI-Align branch necessary?} 
In ROI-AlignPS, the ROI-Align branch extracts the explicitly aligned re-id features, complementary to the implicitly aligned features in the AlignPS branch. To evaluate the effectiveness of this design, we test each branch under different settings. (1) \textit{Separate Training/Test}, i.e., two person search models are separately trained and evaluated. During inference, we also concatenate the features from the ROI-Align model (with RPN) and the AlignPS model. (2) \textit{Joint Training w/o RPN}, where the ROI-Align branch and the AlignPS branch are jointly trained, and both the training and test pipelines follow the structure in Fig.~\ref{subfig-r-2}, i.e., no RPN is employed during training and inference.
(3) \textit{Joint Training w/ RPN}, which denotes the proposed ROI-AlignPS framework, where the training pipeline follows Fig.~\ref{subfig-r-1}, and the test pipeline follows Fig.~\ref{subfig-r-2}. In this case, RPN is only employed in the ROI-Align branch during training. As shown in Table~\ref{tab:rbn}, although the separately trained models both obtain promising results, directly concatenating their re-id features does not yield better performance, which may be because the detection results of the two models are not perfectly aligned. In our joint training setting, in contrast, the branch-level features are well aligned. When the ROI-Align branch is trained without RPN, although the performance of each branch is similar to the separately trained models, the concatenated re-id features are notably improved by 1.3\% and 4.1\% w.r.t. mAP on CUHK-SYSU and PRW, respectively. Moreover, when the ROI-Align branch is trained with RPN, the performance of this branch is further improved, which in turn yields enhanced performance by concatenating the re-id features. These results not only indicate that the jointly training pipeline can better regularize features in the shared layers (i.e., res1-4), but also validate the effectiveness and necessity of the ROI-Align branch, which is complementary to AlignPS.


\textbf{How does the ROI-Align branch influence the efficiency?}
In ROI-AlignPS, the ROI-Align branch inevitably brings additional computational overhead. As shown in Table~\ref{tab:rbn}, in the inference stage, it takes 75 milliseconds (ms) to process an image with both AlignPS and ROI-Align, in comparison to 61 ms for AlignPS only. However, this is still less than the Faster R-CNN based ROI-Align model (83 ms), because our model does not require dense region proposals during inference. Therefore, ROI-AlignPS achieves significant performance improvement and remains highly efficient.

\begin{table}[t]
\small
\centering
\begin{tabular}{p{2.1cm}|p{1.1cm}<{\centering}p{1.1cm}<{\centering}|p{1.1cm}<{\centering}p{1.1cm}<{\centering}}
\hline\thickhline
\rowcolor{mygray} 
  & \multicolumn{2}{c|}{CUHK-SYSU} & \multicolumn{2}{c}{PRW}   \\ \cline{2-5} 
\rowcolor{mygray} 
\multirow{-2}{*}{Mutual Learning}  & mAP & top-1  & mAP  & top-1  \\  \hline \hline 
\textit{None}
   & 95.0     & 95.3       & 50.3   & 84.3   \\
$\mathcal{L}_{mi}$
   & \textbf{95.4}     & \textbf{96.0}      & \textbf{51.6}   & 84.4   \\
$\mathcal{L}_{kl}$  & 95.3   &  95.9      & 50.4   & \textbf{85.3}    \\ 
$\mathcal{L}_{dv}$  & 95.2   & 95.6       & 50.4   & 84.9  \\
$\mathcal{L}_{mi}$+$\mathcal{L}_{kl}$ & 95.4    & 95.8 & 51.1   & 84.0\\
$\mathcal{L}_{mi}$+$\mathcal{L}_{dv}$ & 95.1    & 95.6 & 50.4 & 84.5\\
$\mathcal{L}_{kl}$+$\mathcal{L}_{dv}$ & 95.4    & 95.8 & 50.0 & 84.6\\
$\mathcal{L}_{mi}$+$\mathcal{L}_{kl}$+ $\mathcal{L}_{dv}$ & 95.1    & 95.3 & 51.4 & 85.2\\
\hline
\end{tabular}
\caption{Comparative results on CUHK-SYSU and PRW with different mutual learning loss functions. }
\label{tab:mls}
\end{table}

\begin{figure}[t]
\vspace{-6mm}
\centering
\subfloat[w/o Mutual Learning\label{subfig-tsne1}]{%
   \includegraphics[width=0.485\linewidth]{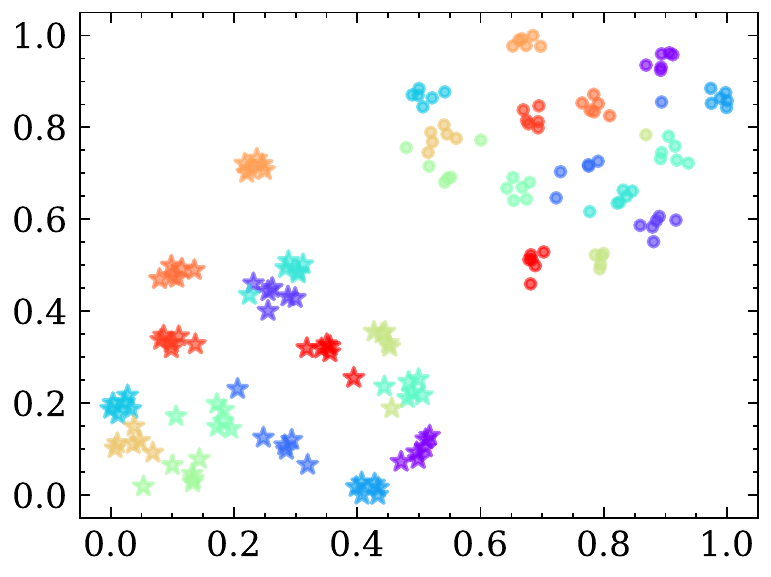}
}
\subfloat[w/ Mutual Information\label{subfig-tsne2}]{%
   \includegraphics[width=0.485\linewidth]{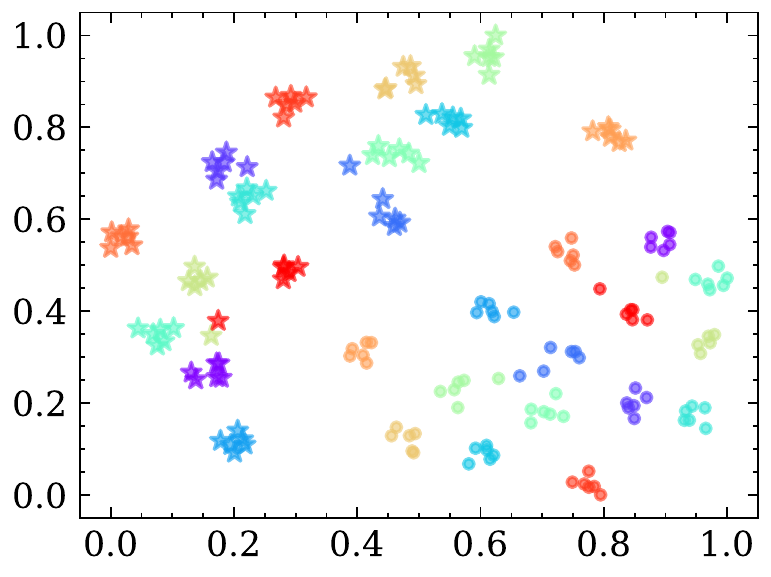}
}
\hfill
\centering
\subfloat[w/ KL-Divergence\label{subfig-tsne3}]{%
   \includegraphics[width=0.485\linewidth]{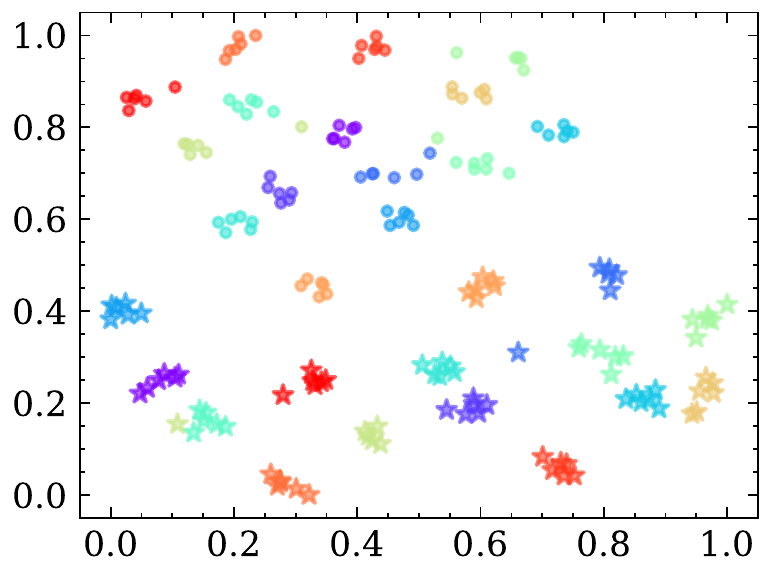}
}
\subfloat[w/ Diversity\label{subfig-tsne4}]{%
   \includegraphics[width=0.485\linewidth]{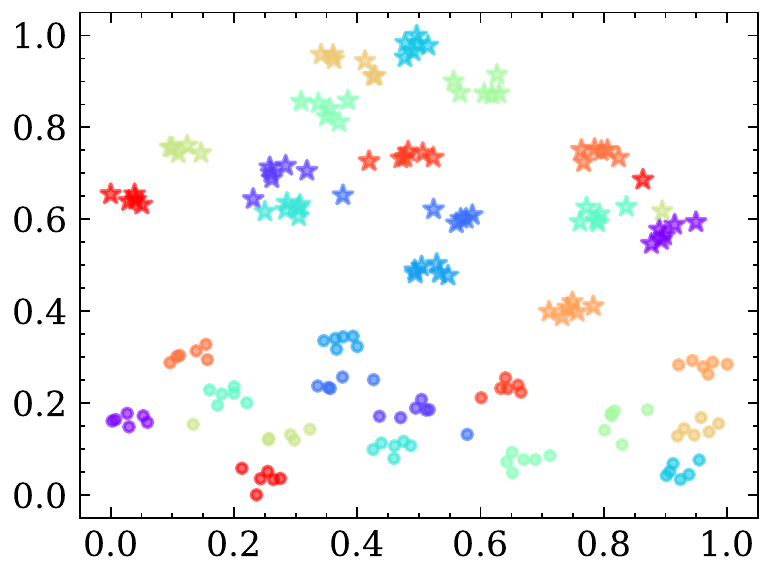}
}
 \caption{The distribution of re-id features from 15 random identities. `*' denotes the re-id features from the ROI-Align branch, and `$\cdot$' denotes the features from AlignPS branch. Different colors represent different identities. }
 \label{fig:tsne}
\end{figure}

\textbf{Are the mutual learning strategies effective?}
As mentioned in Sec.~\ref{sec:mls}, we investigate several mutual learning strategies 
to enhance the representation capabilities of the re-id features in ROI-AlignPS. By applying these strategies, we observe 0.1\%-1.3\% performance gains in mAP on the two datasets, as shown in Table~\ref{tab:mls}. However, combining several strategies together does not yield better performance, because the objectives of these strategies are partially overlapped. Furthermore, we visualize the distribution of re-id features based on t-SNE. As shown in Fig.~\ref{fig:tsne}, the features generated from the two branches are separable, which further validates the complementary property of these two branches. We also observe that the features with mutual learning strategies are more separable, 
where the overall performance with mutual information loss is slightly better than employing other strategies. Therefore, in the following, we denote ROI-AlignPS as the model trained with $\mathcal{L}_{mi}$. 

\begin{table}[t]
\small
\centering
\begin{tabular}{p{0.3cm}|p{2.5cm}|p{0.8cm}<{\centering}p{0.8cm}<{\centering}|p{0.8cm}<{\centering}p{0.8cm}<{\centering}}
\hline\thickhline
\rowcolor{mygray} 
\multicolumn{2}{c|}{}  & \multicolumn{2}{c|}{CUHK-SYSU}                       & \multicolumn{2}{c}{PRW}                             \\ \cline{3-6} 
\rowcolor{mygray} 
\multicolumn{2}{c|}{\multirow{-2}{*}{Methods}}                       & \multicolumn{1}{c}{mAP} & \multicolumn{1}{c|}{top-1} & \multicolumn{1}{c}{mAP} & \multicolumn{1}{c}{top-1} \\ \hline\hline
\multirow{15}{*}{ \rotatebox{90}{one-step}}               & OIM~\cite{DBLP:conf/cvpr/XiaoLWLW17}      & 75.5  & 78.7      & 21.3   & 49.4    \\ 
 & IAN~\cite{DBLP:journals/pr/XiaoXTHWF19}    & 76.3  & 80.1 & 23.0   & 61.9 \\ 
 & NPSM~\cite{DBLP:conf/iccv/LiuFJKZQJY17}        & 77.9  & 81.2 & 24.2   & 53.1 \\
 & RCAA~\cite{DBLP:conf/eccv/ChangHSLYH18}        & 79.3  & 81.3 & -   & - \\
 & CTXG~\cite{DBLP:conf/cvpr/YanZNZXY19} & 84.1  & 86.5 & 33.4   & 73.6 \\
 & QEEPS~\cite{DBLP:conf/cvpr/MunjalATG19} & 88.9  & 89.1 & 37.1   & 76.7 \\
 & HOIM~\cite{DBLP:conf/aaai/ChenZO0S20} &89.7 &90.8 &39.8 &80.4 \\
 & BINet~\cite{DBLP:conf/cvpr/DongZST20a}        & 90.0  & 90.7 & 45.3   & 81.7 \\
 & NAE~\cite{DBLP:conf/cvpr/ChenZYS20}        & 91.5 & 92.4 & 43.3   & 80.9 \\
 & NAE+~\cite{DBLP:conf/cvpr/ChenZYS20}        & 92.1 & 92.9 & 44.0   & 81.1 \\
 & PGA~\cite{Kim_2021_CVPR} & 92.3 & 94.7 &44.2 & 85.2\\
 & DKD~\cite{DBLP:conf/aaai/ZhangWBSY21}        & 93.1 & 94.2 & 50.5   & 87.1 \\
  & SeqNet~\cite{DBLP:conf/aaai/LiM21}        & 93.8 & 94.6 & 46.7   & 83.4 \\
 & SeqNet+CBGM\cite{DBLP:conf/aaai/LiM21}        & 94.8 & 95.7 & 47.6   & \textbf{87.6} \\
 & \textbf{AlignPS} &93.1 & 93.4 &45.9 & 81.9 \\
  & \textbf{ROI-AlignPS} &\textbf{95.4} & \textbf{96.0} &\textbf{\color{gray}51.6} & \textbf{\color{gray}84.4} \\
 \hline \hline
 \multirow{6}{*}{ \rotatebox{90}{two-step}} 
 & DPM+IDE~\cite{DBLP:conf/cvpr/ZhengZSCYT17}        & -  & - & 20.5   & 48.3 \\
 & CNN+MGTS~\cite{DBLP:conf/eccv/ChenZOYT18}        & 83.3  & 83.9 & 32.8   & 72.1 \\
 & CNN+CLSA~\cite{DBLP:conf/eccv/LanZG18}        & 87.2  & 88.5 & 38.7   & 65.0 \\
 & FPN+RDLR~\cite{DBLP:conf/iccv/HanYZTZGS19}        & 93.0 & 94.2 & 42.9  & 70.2 \\
  & IGPN~\cite{DBLP:conf/cvpr/DongZST20}        & 90.3  & 91.4 & 47.2   & 87.0 \\
  & OR~\cite{DBLP:journals/tip/YaoX21} & 92.3  & 93.8 & \textbf{52.3}   & 71.5 \\
  & TCTS~\cite{DBLP:conf/cvpr/WangMCSC20}       & 93.9  & 95.1 & 46.8   & 87.5 \\
  \hline
\end{tabular}
\caption{Comparison with state-of-the-art methods. The upper block lists the results of one-step models, while the lower block shows the results of two-step methods.}
\label{tab:sota}
\end{table}

\begin{figure*}[ht]
\begin{center}
\includegraphics[width=\linewidth]{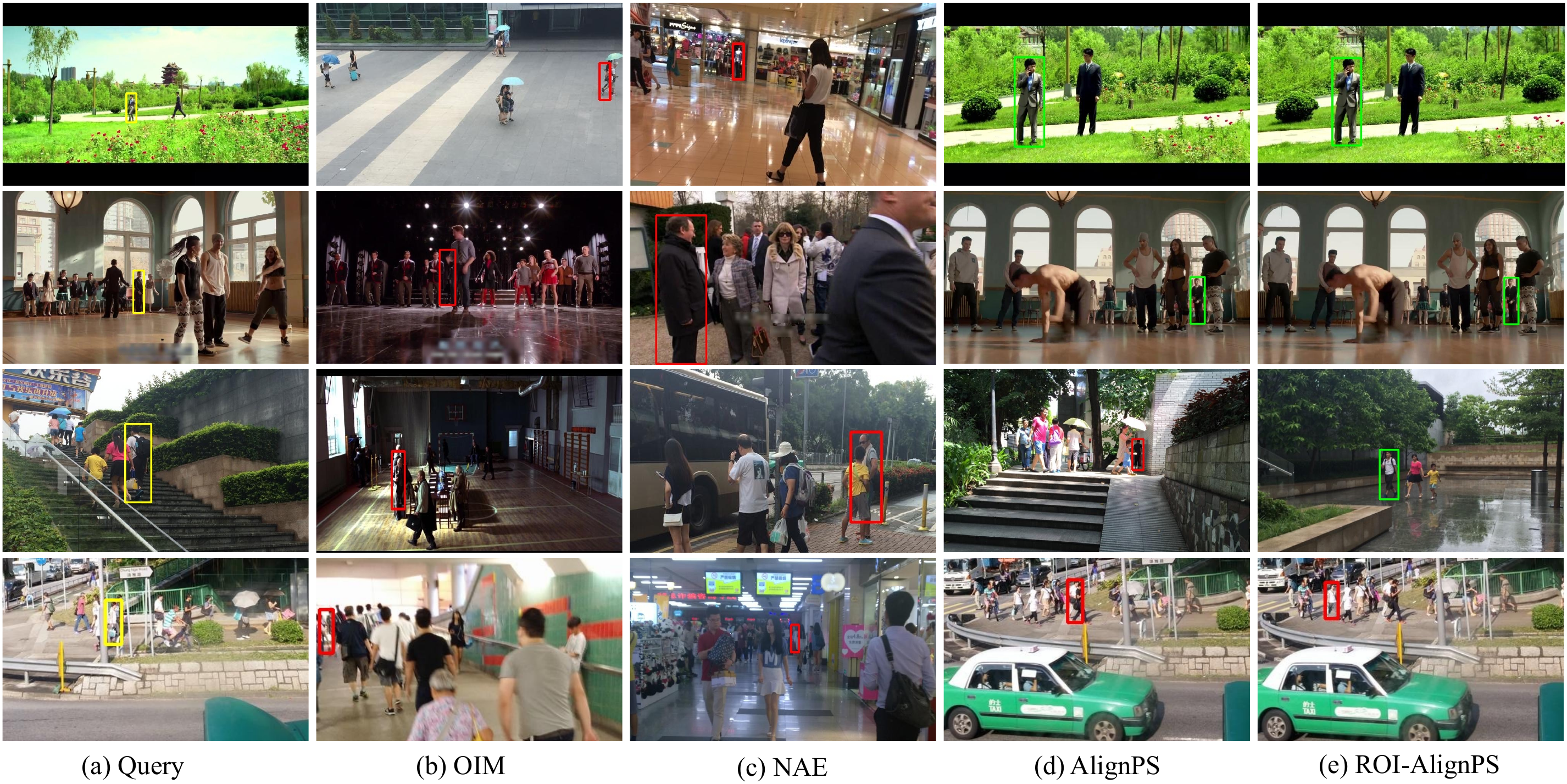}
\end{center}
   \caption{Visualization of some difficult cases.  The yellow bounding boxes denote the queries, while the green and red bounding boxes denote correct and incorrect top-1 matches, respectively.}
\label{fig:vis}
\end{figure*}

\begin{figure}[t]
\centering
\includegraphics[width=\linewidth]{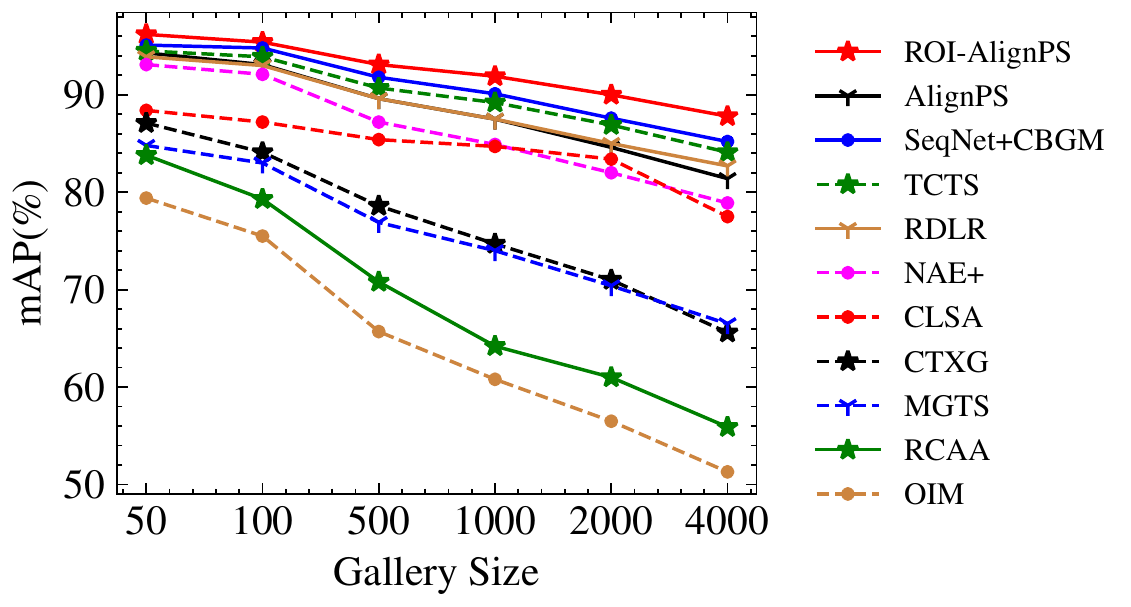}
 \caption{Comparative results on CUHK-SYSU with different gallery sizes. Our models (AlignPS and ROI-AlignPS) are compared with both one-step models and two-step models.}
 \label{fig:gallerysize}
\end{figure}

\subsection{Comparison to State-of-the-Art Methods}\label{sec:cmps}
We compare our framework with state-of-the-art one-step models \cite{DBLP:conf/cvpr/XiaoLWLW17,DBLP:journals/pr/XiaoXTHWF19,DBLP:conf/iccv/LiuFJKZQJY17,DBLP:conf/eccv/ChangHSLYH18,DBLP:conf/aaai/ChenZO0S20,DBLP:conf/cvpr/MunjalATG19,DBLP:conf/cvpr/DongZST20a,DBLP:conf/cvpr/ChenZYS20,Kim_2021_CVPR,DBLP:conf/aaai/ZhangWBSY21,DBLP:conf/aaai/LiM21} and two-step models~\cite{DBLP:conf/eccv/ChenZOYT18,DBLP:conf/eccv/LanZG18,DBLP:conf/iccv/HanYZTZGS19,DBLP:conf/cvpr/DongZST20,DBLP:conf/cvpr/WangMCSC20}. 

\textbf{Results on CUHK-SYSU}.
As shown in Table~\ref{tab:sota}, ROI-AlignPS outperforms all the existing person search models. Notably, ROI-AlignPS outperforms the current best-performing SeqNet~\cite{DBLP:conf/aaai/LiM21} by 0.6\% and 0.3\% in mAP and top-1 accuracy, respectively. Note that SeqNet requires a graph matching strategy as post-processing to achieve its best performance, while ROI-AlignPS does not need such a process.
We also observe from the table that our model outperforms all the two-step models, even though they employ two separate models for detection and re-id. In contrast, our model allows joint inference with a very simple structure, whilst running at a higher speed. 

We visualize the results of AlignPS and ROI-AlignPS w.r.t. mAP with various gallery sizes and compare our model with both one-step and two-step models. Fig.~\ref{fig:gallerysize} illustrates the detailed results, where ROI-AlignPS outperforms all the models by notable margins, in terms of all the gallery sizes.

\textbf{Results on PRW}.
PRW contains less training data; therefore, all the models achieve worse performance on this dataset. Nevertheless, as can be observed from Table~\ref{tab:sota}, ROI-AlignPS still outperforms all the one-step models in terms of mAP. We notice that DKD~\cite{DBLP:conf/aaai/ZhangWBSY21}, PGA~\cite{Kim_2021_CVPR}, IGPN~\cite{DBLP:conf/cvpr/DongZST20}, and TCTS~\cite{DBLP:conf/cvpr/WangMCSC20} achieve higher top-1 accuracy on PRW. These methods also seek to address the feature misalignment issue, but they either resort to feature distillation from the cropped instances~\cite{DBLP:conf/aaai/ZhangWBSY21,DBLP:conf/cvpr/DongZST20,DBLP:conf/cvpr/WangMCSC20}, or apply attention mechanisms~\cite{Kim_2021_CVPR}. Differently, our model efficiently addresses this issue with the proposed AFA module.

\textbf{Efficiency Comparison}.
Since different methods are evaluated with different GPUs, it is difficult to conduct a fair comparison of the efficiency among all the models. Here, we compare our method with OIM\footnote{We test the PyTorch implementation at \url{https://github.com/serend1p1ty/person_search}.} \cite{DBLP:conf/cvpr/XiaoLWLW17}, NAE/NAE+~\cite{DBLP:conf/cvpr/ChenZYS20} and SeqNet~\cite{DBLP:conf/aaai/LiM21} on the same Tesla V100 GPU. All the test images are resized to 1500$\times$900 before being fed to the networks. As shown in Table~\ref{tab:runtime}, our anchor-free AlignPS only takes 61 ms to process an image, which is 27\% and 38\% faster than NAE and NAE+, respectively. Meanwhile, ROI-AlignPS also runs faster than prior state-of-the-art models~\cite{DBLP:conf/cvpr/ChenZYS20,DBLP:conf/aaai/LiM21}. This is because our models are anchor-free during inference, validating the advantage of the proposed framework.

\begin{table}[t]
\small
\centering
\begin{tabular}{p{2cm}p{1cm}<{\centering}p{1cm}<{\centering}p{1cm}<{\centering}p{1.5cm}<{\centering}}
\hline\thickhline
\rowcolor{mygray} 
Methods &Train Anchor &Test Anchor & GPU  &  Time (ms)  \\  \hline \hline  
PGA~\cite{Kim_2021_CVPR} & $\checkmark$& $\checkmark$        &Titan X & 356 \\
DKD~\cite{DBLP:conf/aaai/ZhangWBSY21} &$\checkmark$ &  $\checkmark$      &1080 Ti & 124 \\
OIM \cite{DBLP:conf/cvpr/XiaoLWLW17} &$\checkmark$ &$\checkmark$& V100 & 118 \\
NAE+ \cite{DBLP:conf/cvpr/ChenZYS20} &$\checkmark$ &$\checkmark$ & V100 & 98 \\
NAE \cite{DBLP:conf/cvpr/ChenZYS20} &$\checkmark$ &$\checkmark$ & V100 & 83 \\
SeqNet \cite{DBLP:conf/aaai/LiM21} &$\checkmark$ &$\checkmark$ & V100 & 86 \\
\textbf{AlignPS} &$\times$ &$\times$ & V100 & \textbf{61} \\
\textbf{ROI-AlignPS} &$\checkmark$ & $\times$ & V100 & 75
\\\hline
\end{tabular}
\caption{Runtime comparison of different models. Both AlignPS and ROI-AlignPS are anchor-free during inference.}
\label{tab:runtime}
\vspace{-4mm}
\end{table}

\textbf{Qualitative Results}.
Some qualitative results are illustrated in Fig.~\ref{fig:vis}. We can observe that AlignPS and ROI-AlignPS are more robust in handling occlusions and scale/viewpoint variations, where OIM \cite{DBLP:conf/cvpr/XiaoLWLW17} and NAE \cite{DBLP:conf/cvpr/ChenZYS20} fail. A failure case is illustrated in the last row, where our models fail in distinguishing very tiny objects that share similar appearances. In our future work, we will work towards this direction.

\section{Conclusion}
In this paper, we propose an anchor-free approach to efficiently tackling the task of person search, by developing a person search framework based on an anchor-free detector. We design the aligned feature aggregation module to effectively address the scale, region, and task misalignment issues when accommodating the detector for the person search task. Furthermore, we propose to augment our anchor-free model with an ROI-Align branch, which additionally takes advantage of the anchor-based models. Extensive experiments demonstrate that the proposed framework not only outperforms existing person search methods but also runs at a higher speed.

{\small
\bibliographystyle{IEEEtran}
\bibliography{egbib}
}




\end{document}